\ifdefined\XeTeXversion\else\pdfoutput=1\fi
\documentclass[11pt]{article}

\newif\ifanon

\usepackage[letterpaper,margin=1in,headheight=14pt]{geometry}

\usepackage[table]{xcolor}
\definecolor{mitraink}{RGB}{35,47,62}      %
\definecolor{mitraaccent}{RGB}{255,153,0}  %
\definecolor{mitraaccentdk}{RGB}{179,107,0} %
\definecolor{mitralink}{RGB}{0,90,160}     %

\usepackage[T1]{fontenc}
\usepackage{amsmath,amssymb}
\usepackage{newpxtext,newpxmath}
\usepackage{microtype}
\usepackage{booktabs}              %
\usepackage{array}
\usepackage{multirow}
\usepackage{siunitx}               %
\usepackage{tabularx}
\usepackage{longtable}             %

\usepackage{graphicx}
\usepackage{subcaption}
\graphicspath{{figures/}}
\usepackage{float}
\usepackage{wrapfig}               %
\usepackage{placeins}              %
\usepackage[font=small,labelfont={bf,color=mitraink},skip=6pt]{caption}
\usepackage{tikz}
\usetikzlibrary{positioning,arrows.meta}

\usepackage{titlesec}
\titleformat{\section}{\Large\bfseries\color{mitraink}}{\thesection}{0.8em}{}
\titleformat{\subsection}{\large\bfseries\color{mitraink}}{\thesubsection}{0.8em}{}
\titleformat{\subsubsection}{\normalsize\bfseries\color{mitraink}}{\thesubsubsection}{0.8em}{}
\titleformat{\paragraph}[runin]{\bfseries\color{mitraink}}{\theparagraph}{0em}{}
\titlespacing*{\section}{0pt}{1.7em}{0.8em}
\titlespacing*{\subsection}{0pt}{1.4em}{0.6em}
\titlespacing*{\paragraph}{0pt}{1.2em}{0.8em}

\usepackage{fancyhdr}
\fancypagestyle{plain}{%
  \fancyhf{}\fancyfoot[C]{\small\thepage}}

\usepackage[most]{tcolorbox}

\usepackage{enumitem}
\usepackage[round]{natbib}         %
\usepackage[colorlinks=true,linkcolor=mitralink,citecolor=mitralink,%
            urlcolor=mitralink,pdftitle={Mitra-v2 Technical Report},%
            pdfauthor={\ifanon Anonymous Author(s)\else Amazon\fi}]{hyperref}
\usepackage{cleveref}              %

\newcommand{\mitra}{Mitra-v2}

\newcommand{\reportdate}{3 Sep 2026}

\newcommand{\metalabel}[1]{\makebox[7.5em][l]{\footnotesize\textsc{\textcolor{mitraaccentdk}{#1}}}}

\begin{document}

\thispagestyle{plain}
\vspace*{-30pt}
\begin{center}
{\huge\bfseries\color{mitraink} Mitra-v2 Technical Report\par}
\vspace{9pt}
\ifanon
{\small Anonymous Author(s)\par}
\else
\begin{minipage}{0.9\textwidth}\centering
{\small
  Yefan Tao\textsuperscript{*},\,
  Xiyuan Zhang\textsuperscript{*},\,
  Xinyi Liu\textsuperscript{*},\,
  Boran Han\textsuperscript{*},\,
  Danielle Maddix,\,
  Haoyang Fang,\,
  Zhen Han,\,
  Jiading Gai,\,
  Xuanqing Liu,\,
  Michael Bohlke-Schneider,\,
  Yuyang~(Bernie) Wang,\,
  Gerald Friedland,\,
  Kevan Mah,\,
  Chris Lee,\,
  Chris Kong\textsuperscript{\dag}\par}
\vspace{4pt}
{\small\bfseries\color{mitraaccentdk} Amazon\par}
\vspace{3pt}
{\footnotesize\color{mitraink!75}\textsuperscript{\dag}\,Correspondence: Chris Kong at \texttt{luyankon@amazon.com}
 \quad \textsuperscript{*}\,Equal contribution.\par}
\end{minipage}
\fi
\end{center}
\vspace{4pt}

\begin{tcolorbox}[enhanced, sharp corners, boxrule=0pt, frame hidden,
  borderline west={3pt}{0pt}{mitraaccent},
  colback=mitraaccent!3, left=13pt, right=10pt, top=6pt, bottom=6pt]
{\footnotesize\textsc{\textcolor{mitraaccentdk}{Abstract}}\par}
\vspace{2.5pt}
{\small\setlength{\parindent}{0pt}\setlength{\parskip}{4pt}%
We introduce \textbf{\mitra}, a tabular foundation model that delivers state-of-the-art performance on real-world classification and regression problems, from credit-risk scoring and clinical prediction to equipment-failure detection and house-price estimation. \mitra{} is trained only on synthetic data, with a pretraining distribution that is much larger and more diverse than Mitra-v1's. Built on a small 2D Transformer backbone, \mitra{} supports longer contexts and larger feature spaces. Improved optimization lets it learn from this larger task distribution. We evaluate \mitra{} on the TabArena and TALENT benchmarks, comprising more than 300 real-world datasets under two evaluation protocols. On the full TabArena benchmark, \mitra{} delivers state-of-the-art performance at the level of the industry-scale TabFM and EXAONE Tabular models, while surpassing TabPFN-3 by a wide margin in both classification and regression. \mitra{} matches the 1.6B-parameter TabFM with only 5\% of its size (77M parameters), delivering frontier performance at a fraction of the cost. On TALENT, \mitra{} remains among the leading models, clearly outperforming TabPFN-3 and TabICLv2. It also ranks first on classification tasks with more than ten classes, even though it was pretrained only on tasks with at most ten classes. These results make \mitra{} one of the strongest and most broadly applicable open tabular foundation models released to date. We release the model weights, the inference and fine-tuning code, and our evaluation results under the Apache-2.0 license.

\par}
\end{tcolorbox}
\vspace{5pt}
{\small\setlength{\parindent}{0pt}\hspace{13pt}%
  \metalabel{Date}\reportdate\\[2.5pt]
  \hspace*{13pt}\metalabel{Model}%
  \href{https://huggingface.co/autogluon/mitra-classifier-2}{\texttt{autogluon/mitra-classifier-2}},\;
  \href{https://huggingface.co/autogluon/mitra-regressor-2}{\texttt{autogluon/mitra-regressor-2}}\\[2.5pt]
  \hspace*{13pt}\metalabel{Code \& results}%
  \href{https://huggingface.co/autogluon/mitra-finetune}{\texttt{autogluon/mitra-finetune}}\par}
\vspace{2pt}

{\centering
\begin{minipage}{\textwidth}
  \centering
  \includegraphics[width=\linewidth]{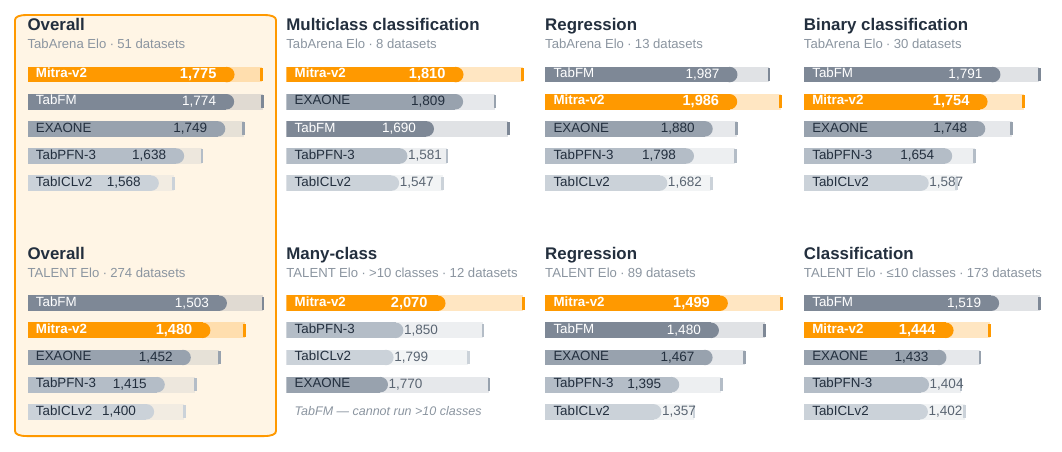}
  \captionof{figure}{\mitra{} across the TabArena boards (top; one-hour
  protocol) and the TALENT boards (bottom; 274 datasets, train-only matched
  context policy). Panels are ranked by Elo on independent truncated scales; the
  shaded band beyond each bar is its asymmetric 95\% bootstrap interval.}
  \label{fig:hero}
\end{minipage}
\par}
\vspace{8pt}

\clearpage
\setcounter{tocdepth}{2}
{\small\tableofcontents}
\clearpage

\section{Introduction}
\label{sec:intro}

Tabular foundation models reframe supervised tabular learning as amortized
inference. A single model is pretrained across many prediction problems; at
deployment it conditions on the labeled rows of a new dataset rather than being
fit from scratch for each task. The paradigm was introduced by
TabPFN \citep{hollmann2022tabpfn} and carried forward by the same group. Their
follow-up TabPFNv2 \citep{hollmann2025tabpfn} drew wide attention for its
accuracy on small tabular data. The later TabPFN-2.5 and TabPFN-3
technical reports advanced the state of the art further
\citep{grinsztajn2025tabpfn25,grinsztajn2026tabpfn3}. It has since grown into an
active family of models with varying architecture, pretraining data, and target dataset scale: in-context learners built for larger tables
\citep{qu2025tabicl,qu2026tabiclv2}, semantics- and text-aware models trained on
real data \citep{arazi2025tabstar,spinaci2025contexttab}, large generalist
structured-data models \citep{zhang2025limix}, and recent industry-scale releases
such as Google's TabFM \citep{google2026tabfm} and LG AI Research's EXAONE Tabular
\citep{lgai2026exaonetabular}. Across these models, the
paradigm shifts part of the modeling burden from downstream hyperparameter search
to three design choices: the pretraining task distribution, the architecture used
to consume a table, and the adaptation protocol used at deployment.

Mitra-v1 \citep{zhang2025mitra} focused on the first of these ingredients. It
showed that a mixture of structural causal models and complementary tree-based
generators improved over relying on a single synthetic prior, and that the same
mixture transferred across row-wise and element-wise architectures. \mitra{} asks a
practical follow-up question: \textbf{how far can this recipe be pushed by
expanding the range of tasks and the internal diversity of the synthetic prior
without making the backbone deeper?}

\mitra{} retains the small 2D attention backbone of Mitra-v1 while substantially expanding the scale and diversity of its synthetic pretraining distribution. Improvements in optimization and downstream adaptation enable the model to handle longer contexts, wider tables, and more complex prediction tasks. Lightweight extensions further allow the same model to operate beyond its native feature and class limits without additional training. These advances are detailed in \Cref{sec:model}.

Across the full TabArena and TALENT benchmarks, \mitra{} delivers state-of-the-art performance (\Cref{fig:hero}), matching the industry-scale TabFM and EXAONE Tabular models while surpassing TabPFN-3 by a wide margin in both classification and regression. The improvements over Mitra-v1 hold across most task groups and are largest in regression. Beyond controlled benchmarks, the Mitra family has been independently adopted and evaluated in published studies across healthcare, finance, energy, transportation, and manufacturing. \Cref{app:usecases} lists these real-world use cases.

\mitra{} delivers this level with a small model. Its released models have
77M parameters (75.7M for classification, 76.7M for regression), on the same
12-layer backbone as Mitra-v1. TabFM and EXAONE Tabular are the two models statistically level with \mitra{}
at the top of both benchmarks; TabFM has 1.6B parameters, roughly 21 times more. EXAONE
Tabular and the remaining foundation models in the comparison (TabICLv2, TabPFN-3) are smaller than \mitra{}, in the 20M to 60M range,
so \mitra{} sits between them and TabFM in size while matching the largest
model's accuracy. Parameter count is only one axis of cost: the default
\mitra{} deployment fine-tunes and bags eight copies of the model, so its
wall-clock is well above that of zero-shot forward-pass models
(\Cref{sec:efficiency}). The point is that its accuracy is not bought with
model size.

\subsection{From Mitra-v1 to \mitra}
\label{sec:v1tov2}

\Cref{tab:v1v2} summarizes the main changes. The Mitra-v1 and \mitra{} Elo values are
intentionally omitted because the original Mitra-v1 paper and this report use different
task sets, method pools, scorer revisions, and evaluation protocols. Elo is
pool-dependent and cannot be compared across those settings.

\begin{figure*}[!tb]
    \centering
    \includegraphics[width=1.0\linewidth]{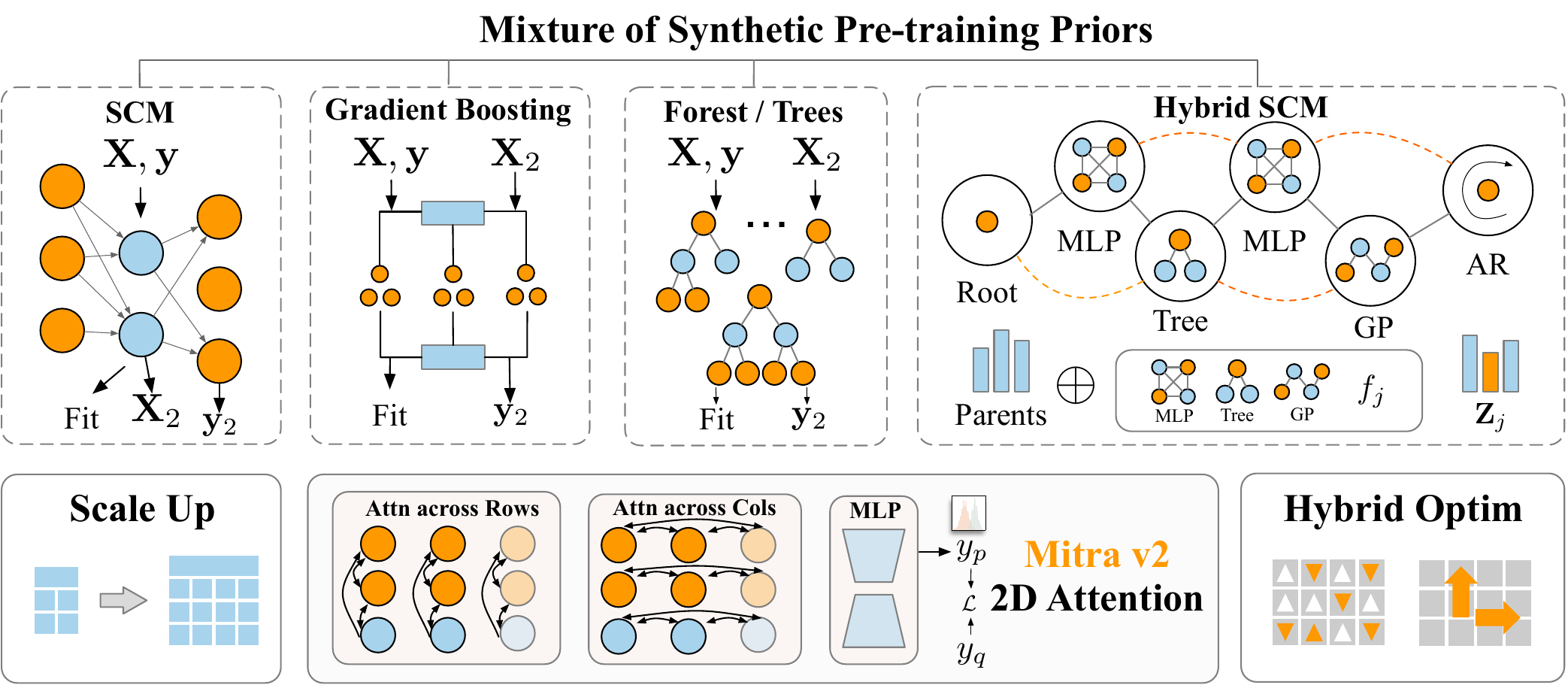}
    \caption{Overview of \mitra{}. \textbf{Top:} the mixture of synthetic pretraining priors: classical SCM, gradient-boosting, and forest/tree generators alongside the new Hybrid SCM prior, which composes neural, tree, Gaussian-process, and autoregressive mechanisms inside a single causal graph. \textbf{Bottom:} the scaled-up pretraining envelope, the element-wise 2D-attention backbone (attention across rows and across columns) carried over from Mitra-v1, and the hybrid optimizer.}
    \label{fig:main}
\end{figure*}

\begin{table}[!ht]
\centering
\caption{Mitra-v1 to \mitra{} changes. Only components that differ are
listed; both versions share the same 12-layer 2D attention backbone (hidden
width 512, four attention heads) and the at-most-ten-class pretraining head.}
\label{tab:v1v2}
\begin{tabularx}{\linewidth}{l >{\raggedleft\arraybackslash}X >{\raggedleft\arraybackslash}X}
\toprule
Component & Mitra-v1 & \mitra{} \\
\midrule
Parameters                            & cls 75.7M / reg 75.7M & cls 75.7M / reg 76.7M \\
Maximum support rows in pretraining   & 512 & 5{,}120 \\
Query rows in pretraining             & 128 & 1{,}280 \\
Maximum features in pretraining       & 16 & 50 \\
Prior mixture                         & SCM + tree priors & SCM + Hybrid SCM + tree priors \\
Training steps                        & 22{,}000 & cls $\approx$14{,}000 / reg $\approx$14{,}000${+}$3{,}000 FT \\
Nominal tasks                         & 45M & cls 28.7M / reg 34.8M \\
Training hardware                     & 8$\times$A100 40GB & 4 nodes $\times$ 8 H200 \\
Main optimizer                        & AdamW & Muon \\
\bottomrule
\end{tabularx}
\end{table}

The architecture changes only modestly in parameter count. The central \mitra{}
hypothesis is instead that a wider synthetic task distribution can improve
downstream transfer:
\begin{itemize}[nosep]
  \item \textbf{Wider task range.} Support tables are sampled between 160 and 5{,}120 rows with a query size of 1{,}280, and each task contains 1 to 50 features.
  \item \textbf{A more diverse prior.} We introduce the Hybrid SCM, a new synthetic-data prior that can place qualitatively different mechanisms in successive nodes of one causal graph, adding within-task structural heterogeneity. It is detailed in \Cref{sec:hybridscm}.
  \item \textbf{Geometry-aware composite optimization.} We replace the uniform AdamW recipe with a parameter-structured optimizer stack to better align updates with parameter geometry and improve optimization efficiency at scale.

\end{itemize}

\FloatBarrier
\section{Model and synthetic pretraining}
\label{sec:model}

\subsection{Tab2D backbone}
\label{sec:backbone}

\mitra{} uses an element-wise 2D Transformer with 12 layers, hidden dimension
512, and four attention heads. Each scalar cell value is embedded independently to
the hidden dimension, support labels are embedded
and added to their rows, and query positions whose label is unknown receive a
learned mask token. Every layer applies attention twice, once across examples (the
row axis) and once across features (the column axis), with each attention sublayer
followed by its own feed-forward block of inner dimension 2{,}048. This factorized
``2D'' attention preserves the structural bias of Mitra-v1: the model exchanges
information both among examples for a feature and among features for an example.
The backbone uses no positional encoding, so its predictions are invariant to the
ordering of both rows and columns.

\mitra{} has roughly 75.7M (classification) and 76.7M (regression) trainable parameters. Keeping depth at 12 follows the
v1 observation that gains from additional depth began to saturate beyond this
point \citep{zhang2025mitra}.
\mitra{} therefore relies on data and optimization scaling rather than on
additional depth.

For a synthetic task with support set
$\mathcal{D}_{s}=\{(x_i,y_i)\}_{i=1}^{n_s}$ and query features $X_q$, the model
predicts the masked query labels and minimizes cross-entropy:
\begin{equation}
\mathcal{L}(\theta)
=
\mathbb{E}_{\mathcal{D}\sim \mathcal{G}}
\left[
-\sum_{j=1}^{n_q}
\log p_\theta(y_{q,j}\mid X_s,y_s,X_q)
\right],
\end{equation}
where $\mathcal{G}$ is the mixture of synthetic task generators described below.

\paragraph{Task-specific heads.} The 2D attention backbone is shared across problem
types. Only the label embedding and the output head change with the task, so the
classification and regression checkpoints differ in size only at the head
(\Cref{tab:v1v2}). Classification embeds up to 10 categorical labels and applies a
linear head over class logits. Targets of more than 10 classes are handled
without retraining by the hierarchical wrapper of \Cref{sec:beyondcaps}. This
classification head is unchanged from Mitra-v1, so the Mitra-v1 and \mitra{}
classification checkpoints have the same architecture and parameter count.

Regression is cast as classification over the target range. The target is
discretized into 1{,}000 bins, each support target is embedded by its bin, and the
head predicts a distribution over those bins. The cross-entropy objective above
then applies unchanged, now over bins rather than classes. A scalar prediction
is recovered as the mean of the predicted bin distribution. This 1{,}000-bin
distributional head replaces the single-output, mean-squared-error regression head
of Mitra-v1 and adds roughly one million parameters. It is the only architectural
difference between the two regression versions.

\subsection{Outer prior mixture}
\label{sec:priormix}

For each pretraining task, a single generator is drawn from a top-level
categorical mixture. We call this the \emph{outer} mixture because it sits above
the within-task mechanism mixing of the Hybrid SCM (\Cref{sec:hybridscm}). The
outer draw selects one generator per table, whereas the Hybrid SCM combines
several mechanisms inside a single table. \Cref{tab:priormix} lists the
configured probabilities used in \mitra{}:

\begin{table}[h]
\centering
\caption{Configured outer prior mixture of \mitra{} (classification). Hybrid SCM
is the generator newly introduced in \mitra{}.}
\label{tab:priormix}
\begin{tabular}{lr}
\toprule
Generator & Weight \\
\midrule
Base TabPFN/SCM & 0.35 \\
Hybrid SCM (new) & 0.35 \\
Decision tree & 0.06 \\
ExtraTrees & 0.06 \\
Gradient boosting & 0.06 \\
Random forest & 0.06 \\
Directly sampled random forest & 0.06 \\
\midrule
Total & 1.00 \\
\bottomrule
\end{tabular}
\end{table}

Hybrid SCM enters with the same weight as the base SCM, so the two causal
generators together hold most of the probability mass. The five tree-based
priors still give direct coverage of tree-generated decision boundaries.

These weights are chosen manually rather than learned, and are not necessarily
optimal. Recent work adapts the synthetic generator distribution during
pretraining to emphasize harder tasks \citep{peroni2025robust}, which suggests the
mixture is a design choice worth tuning. In our own experiments, however, varying
these weights did not produce a large change in downstream accuracy
(\Cref{app:negative}). We therefore keep the hand-chosen values above and leave a
systematic optimization of the mixture to future work.

\subsection{Hybrid SCM}
\label{sec:hybridscm}

We implement a variant of the
Hybrid Structural Causal Model (Hybrid SCM) data generator introduced by the O-prior framework \citep{bouadi2026shaping}. It composes multiple mechanism families $f$ (random MLPs, decision trees, ExtraTrees, random forests, 1D convolutions, Gaussian processes via Random Fourier Features, and Vector AutoRegression) within a single causal DAG, which widens the range of functional relationships the prior can express.

The Hybrid SCM panel of \Cref{fig:main} sketches this construction at a high
level. The process first samples a random DAG with three to seven nodes, where node $X_0$ contains root causes, each node $X_i$ ($i \geq 1$) receives node $X_{i-1}$ as a parent, and each earlier node $X_j$
($j < i-1$) is independently added as a skip-connection parent with probability $0.4$; we write $\mathrm{pa}(i)$ for the resulting parent set of node $i$.

Our variant differs from O-prior in two respects. First, while O-prior uses
scalar-valued nodes (output dimension $1$ for all nodes), our implementation uses
vector-valued nodes: each node $i$ independently samples an output dimensionality
$d_i$ from a range and produces an output $h_i \in \mathbb{R}^{d_i}$. Second, our
node-computation process differs. For each non-root node $i$, we concatenate its
parents' outputs,
$c_i = \mathrm{concat}\big(\{h_j\}_{j \in \mathrm{pa}(i)}\big) \in
\mathbb{R}^{\sum_{j \in \mathrm{pa}(i)} d_j}$,
and apply a randomly selected mechanism $f_i$ to produce a transformed output
$\mathrm{raw}_i = f_i(c_i) \in \mathbb{R}^{d_i}$. When a node has more than one parent
($|\mathrm{pa}(i)| > 1$), we additionally project each parent output to the
mechanism's output dimensionality through independent random linear maps
$W_{ij} \in \mathbb{R}^{d_i \times d_j}$, $p_{ij} = W_{ij}\,h_j$, and aggregate the
mechanism output together with all projected parent outputs,
\begin{equation}
  h_i \;=\; \phi_i\!\big(\{\mathrm{raw}_i\} \cup \{p_{ij}\}_{j \in \mathrm{pa}(i)}\big),
\end{equation}
using a randomly selected operator $\phi_i$ (mean, softmax-weighted sum, MLP projection, elementwise product, or maximum). This creates residual-connection-like paths where parent information flows both through the nonlinear mechanism and through direct linear projections.

Node outputs undergo per-node standardization with random feature-importance scaling and L2-normalization. Final features $\mathbf{X}$ and target $\mathbf{y}$ are selected from the concatenated node outputs using diversity-aware strategies (k-means, farthest-point sampling, or random selection). For classification, the selected target is quantile-bucketed into class labels. We apply categorical feature transformations to convert a subset of continuous features to categorical in all generated datasets.

Figures~\ref{fig:umap-low} and~\ref{fig:umap-high} (Appendix \ref{app:2D_visualization}) show 2D UMAP projections (after PCA whitening) of Hybrid SCM datasets for the 1--16 feature range shared with Mitra-v1 and the higher 17--50 range unique to our variant, respectively. At matched dimensionality (Figure~\ref{fig:umap-low}), Hybrid SCM produces clearly varied dataset geometry.

\subsection{Task shapes and preprocessing}
\label{sec:taskshapes}

The classification configuration samples 160--5{,}120 support rows; 1{,}280 query
rows under the fixed-cap policy; 1--50 features; up to 10 classes; and numerical
and synthesized categorical features. The number of support rows is sampled
uniformly over this range; we also experimented with log-uniform sampling but
observed little difference in downstream accuracy (\Cref{app:negative}).

Synthetic tables are generated online. The preprocessor is fitted to each support
set, then applies the same transformation to its query rows. Class and feature
order are shuffled during pretraining, and the input may be mirrored along
numerical axes. The selected run does not enable the optional quantile transformer
or feature-count scaling.

\subsection{Optimization and distributed training}
\label{sec:optim}

The selected release checkpoint is at roughly 14{,}000 optimizer updates; the
regression checkpoint is fine-tuned for a further 3{,}000 steps. The runtime uses four nodes with eight GPUs each (32
distributed ranks); bfloat16 autocast; four tasks per GPU per microbatch; 16
gradient-accumulation microbatches; a nominal global effective batch of
$4\times32\times16=2{,}048$ tasks; Muon on the two-dimensional weight matrices
other than the prediction head, via the Gram Newton-Schulz implementation
\citep{dao2026gramns}; AdamW for the remaining parameters; base learning rate
$8\times10^{-4}$; cosine decay with 500 warmup steps; and a maximum gradient norm
of 10.

At 2{,}048 tasks per update, the roughly 14{,}000-update checkpoint corresponds to
about \textbf{28.7 million nominal synthetic tasks}. We use ``nominal'' because an
exact count would depend on training-time metadata we did not fully log.

\subsection{Handling tables beyond the pretraining caps}
\label{sec:beyondcaps}

\begin{figure}[t]
\centering
\begin{minipage}[t]{0.48\linewidth}
  \centering
  \includegraphics[width=0.98\linewidth]{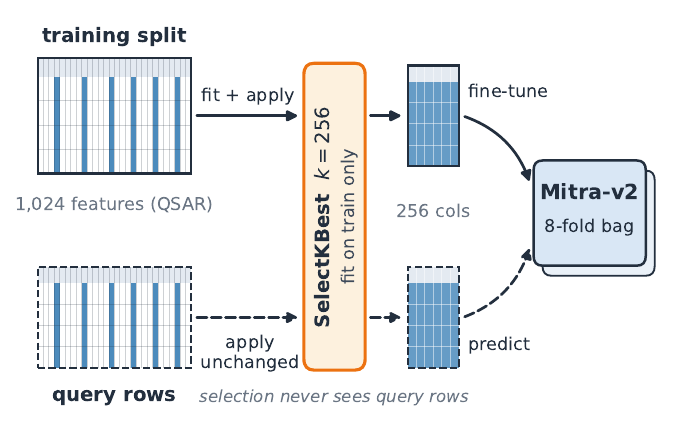}
  \\[3pt]
  {\small (a) Wide tables}
\end{minipage}
\hfill
\begin{minipage}[t]{0.48\linewidth}
  \centering
  \includegraphics[width=0.98\linewidth]{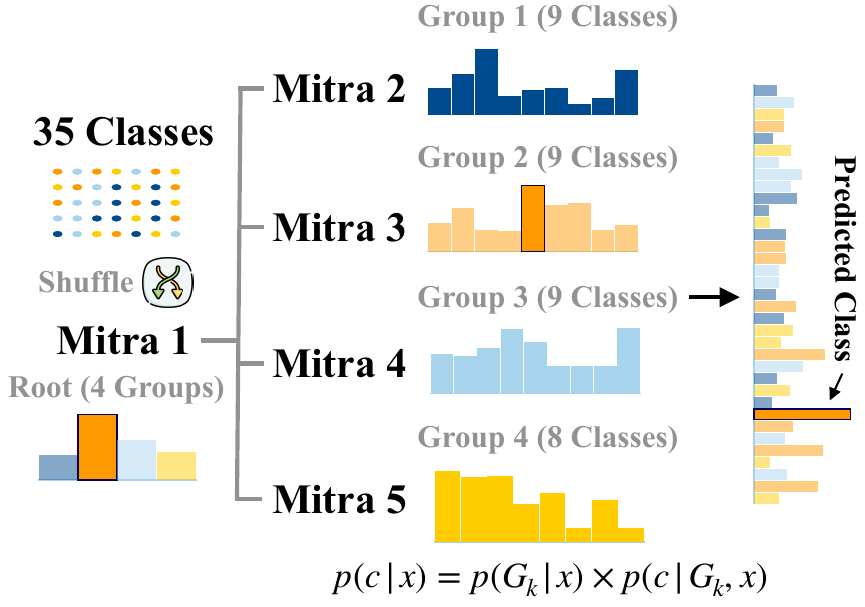}
  \\[3pt]
  {\small (b) Many classes}
\end{minipage}
\caption{Two deployment-time wrappers that let the frozen checkpoint handle tables
beyond the pretraining caps, both reusing the same checkpoint with no additional
pretraining. (a) Wide tables: train-only feature reduction shrinks the column set
to a fixed budget before fine-tuning and bagging. (b) Many classes: a balanced
hierarchical decomposition routes a query through a shallow tree of
at-most-10-class Mitra nodes.}
\label{fig:beyondcaps}
\end{figure}

Pretraining fixes hard caps on table shape: at most 50 features and at most 10
classes (\Cref{sec:taskshapes}). Two deployment-time wrappers, summarized in
\Cref{fig:beyondcaps}, let the same frozen checkpoint be applied to datasets that
exceed these caps, without any additional pretraining.

\paragraph{Wide tables: train-only feature reduction.}
For tables with many more columns than the pretraining ceiling, the downstream
pipeline first reduces the feature set to a fixed budget of 256. For
classification, a \texttt{SelectKBest} filter with $k=256$ (F-statistic scoring)
is fit on the training split only and then applied unchanged to the query rows,
so no query-set information leaks into the selection. It is dtype-gated to fire
only on predominantly continuous wide tables. For regression, the same budget is
instead met by a train-only 256-component truncated-SVD projection fit on the
training split alone (\Cref{sec:regression}). Fine-tuning and bagging then proceed
on the reduced table. Both suites contain such wide tables (hundreds to thousands
of features). The 1{,}024-feature QSAR task is the only TabArena regression table above the
budget, so it is the one reduced by the truncated-SVD projection.

\paragraph{Many classes: balanced hierarchical decomposition.}
The classification head emits at most 10 logits, so datasets with more than 10
classes are handled by decomposing the label set into a shallow tree of
at-most-10-class subproblems. For a task with $C>10$ classes: (i) the classes are
permuted under a fixed seed and split into $g=\min(\lceil C/10\rceil,\,10)$
balanced groups of at most 10 classes each; (ii) a \emph{root} Mitra node routes a
query to one of the $g$ groups; and (iii) one \emph{leaf} Mitra node per group
predicts the class within that group. The probability of an original class is the
product of its routing probability and its within-group probability; these
products are remapped to the original class order, normalized, and averaged across
the bagged trees. When a group still holds more than 10 classes ($C>100$), the
leaf node is itself replaced by a hierarchical classifier and the construction
recurses. The per-tree seed makes each bagged member use a different
class-to-group assignment, which diversifies the ensemble. This wrapper
drives the many-class results: on TALENT's twelve many-class datasets it
places \mitra{} first in the entire comparison pool under both context
policies (\Cref{sec:talent}), even though the backbone never saw more than
ten classes during pretraining.
\section{Evaluation on TabArena}
\label{sec:experiments}

\subsection{Setup and protocol}
\label{sec:setup}

\paragraph{Benchmark and protocol.}
We evaluate on TabArena, a continuously maintained benchmark for tabular machine
learning \citep{erickson2025tabarena}, under its full multi-split protocol rather
than a single split. The classification view has 38 datasets (30 binary tasks
scored by ROC-AUC error and 8 multiclass tasks scored by log-loss), evaluated over
594 evaluation units, where a unit is a single train/test split: every dataset uses
three-fold cross-validation, repeated ten times on the 12 smaller datasets and three
times on the 26 larger ones ($12\times30+26\times9=594$).
The regression view has 13 datasets over 222 units ($5\times30+8\times9=222$ under the
same rule), and the combined board covers all 51 datasets over 816 units. Results are scored against the latest public
TabArena reference leaderboard, whose comparison pool includes strong tuned
baselines such as AutoGluon \citep{erickson2020autogluon}, pre-tuned MLPs and
boosted trees \citep{holzmueller2024realmlp}, parameter-efficient ensembles such as
TabM \citep{gorishniy2025tabm}, and the current tabular foundation models (TabPFN-3,
TabPFN-2.6, TabICLv2, EXAONE Tabular, and TabFM).

\paragraph{Coverage and imputation.}
\mitra{} runs natively on every
dataset in the benchmark. Its operating range covers tables wider than the
50-feature pretraining cap (reduced to a 256-feature budget by train-only feature
reduction), targets with more than ten classes (a balanced hierarchical
decomposition), large in-context support sets (capped at up to 32{,}768 rows at prediction
time), and regression as well as classification (the 1{,}000-bin distributional head
of \Cref{sec:backbone}). All of these reuse the same frozen checkpoint with no
additional pretraining (\Cref{sec:beyondcaps}). \mitra{} therefore has a real result
on every unit, so its imputation rate is 0\%. Mitra-v1, evaluated in the same pool,
is still imputed on 32\% of classification datasets and 46\% of regression datasets, and
\mitra{}'s wider operating range closes that gap. Every method in the
main-text leaderboards below also has 0\% imputed cells, so those rankings are not an
artifact of imputed scores. The complete pools, including partially imputed methods
with their imputation fractions, are given in \Cref{app:boards}.

\paragraph{Time budget and display policy.}
All \mitra{} results in this report are produced under TabArena's one-hour
per-unit budget: every evaluation unit (one fit of the bagged system plus its
prediction) runs with a 3{,}600\,s fit time limit, enforced by a fixed
250\,s fine-tuning budget per bagged child and by truncating the bag at the time
limit when a large dataset cannot complete all folds. The limit governs the fit;
because the harness does not time prediction separately, the recorded
end-to-end unit time also includes bagged prediction and exceeds 3{,}600\,s on
the two widest of the large binary tables (APSFailure and kddcup09\_appetency, about 1.3 to
1.4 hours per unit), while the fit itself is truncated at the limit by
construction. We flag one asymmetry in
the comparison pool: the TabFM (default) artifacts on the
reference leaderboard include per-dataset fits whose fit time alone exceeds this budget (up to
5{,}886\,s), so its rating reflects a larger compute envelope than ours. We also
exclude the TabFM+ entry from the main-text boards: it is a multi-view
foundation-model ensemble that TabArena lists in its four-hour system class, not a single-model
default. It remains in the complete pools of \Cref{app:boards}, and all Elo
ratings are computed within the full frozen pool that contains it.

\paragraph{Downstream Mitra configuration.}
The evaluated method is AutoGluon's bagged Mitra fine-tuning system, run through
the released \texttt{mitra-finetune} package (whose fine-tuning defaults differ
from AutoGluon's stock settings) with the \mitra{}
weights injected into the model wrapper. It performs 50 fine-tuning steps by
default, capped at a 250\,s fine-tuning budget per bagged child under the
benchmark time limit; one Mitra estimator in each internal fold; and eight-fold
AutoGluon bagging. The downstream fine-tuning learning rate is $10^{-5}$, reduced
to $3\times10^{-6}$ on binary tasks whose training table has at most 16{,}384
rows, with a 10-step warmup and weight decay 0.3. The in-context support set is capped separately for the two phases:
fine-tuning sees at most 16{,}384 rows on classification and 20{,}480 on
regression, while prediction draws up to 16{,}384 rows on binary classification
and 32{,}768 on multiclass classification and regression (with an automatic
halving ratchet under GPU memory pressure). The lower binary cap is a
time-budget choice rather than an accuracy optimum: on the largest binary tables
a wider prediction context slowed the bag enough to trigger truncation under the
one-hour limit, which cost more than the added context gained. On binary tasks the prediction-time
support subsample is class-balanced rather than uniformly random. After the
bagged fit, each bag child predicts with its full outer training fold as
in-context support, that is, its fit fold plus its own held-out fold (``heldout in
support''); the held-out labels are used only as fine-tuning validation and as
prediction-time support rows, so no test information is involved. Wide tables are
additionally reduced by the train-only feature-budget rule (top-256 selection for
classification, truncated SVD for regression; \Cref{sec:beyondcaps}). The same configuration is used for every
dataset; nothing in this report uses per-dataset selection of any kind
(\Cref{app:negative}). Consequently, labels such as ``default'' in the
leaderboard mean default \textbf{fine-tuned and bagged} Mitra. They do not mean
zero-shot inference. This distinction is especially important when comparing against
TabPFN-3 default, whose report emphasizes forward-pass performance
\citep{grinsztajn2026tabpfn3}.

\paragraph{Metrics and uncertainty.}
TabArena converts task results into pairwise method outcomes and fits an Elo-style
Bradley--Terry ranking. Elo is a relative, pool-dependent summary: adding or
removing methods can change every rating, so we report all Elo comparisons within
the same frozen reference pool. The leaderboard includes asymmetric 95\% bootstrap
intervals from 200 rounds; these describe sampling uncertainty under the benchmark
bootstrap and do not account for implementation choices. Because these intervals are wide relative to the gaps
between the leading methods, we read close Elo differences as ties rather than as
decisive separations. The front-page overview (\Cref{fig:hero}) additionally
splits the classification board into its binary (30-dataset) and multiclass
(8-dataset) sub-boards. These are TabArena's own comparison run on each task
subset within the same frozen pool and anchoring, so their Elo values are not
comparable across sub-boards or with the full board.

\subsection{Overall standing}
\label{sec:overall}

Across the full 51-dataset, 816-unit board, \mitra{} delivers the state-of-the-art
level set by TabFM and EXAONE Tabular: at 1{,}774.6 Elo it sits at the top of the
single-model field, statistically level with TabFM (whose point estimate it edges
by about 1 Elo) and roughly 25 Elo above EXAONE Tabular. It surpasses TabPFN-3 by
roughly 137 Elo.

\paragraph{Rank-one counts.}
The leaderboard's rank-one count is the average number of datasets on which a
method places first. It is informative because a high rating can arise in two
different ways. A four-hour AutoML portfolio fields many models
per dataset, so it is rarely bad and also rarely the best, and its rating
comes from consistently placing near the top. A single frozen model runs one configuration everywhere, so its rating comes from staying close to the leader on most datasets or from placing first. \mitra{}'s rating is of the second sort: it
places first on about 9 of the 51 datasets, more than twice EXAONE Tabular's 3.6
and more than all four AutoGluon portfolios combined (5.7). The two profiles suit different use cases: the portfolio is the safe default,
while \mitra{} is more often the single best-performing method on a given dataset.

\paragraph{What sits above \mitra{}.}
Excluding the four-hour TabFM+ ensemble (\Cref{sec:setup}), the
only entry above \mitra{} on the Elo board is a single four-hour AutoGluon
AutoML portfolio (the noncommercial 1.6 preset), which is not a
single-configuration foundation model running within the benchmark time limit.
TabFM, whose leaderboard artifacts include fits well beyond the one-hour budget
(\Cref{sec:setup}), is statistically level with \mitra{} just below it, and the
AutoGluon 1.6 extreme, 1.5, and 1.4 presets all rank lower. \Cref{fig:hero}
(Overall panel) visualizes the top of this board and \Cref{tab:overall} lists the
leaders.

\paragraph{Margins, not only wins.}
Elo, average rank, and rank-one counts all reduce each task to a win or a
loss; the gap-to-best column of \Cref{tab:overall} keeps the margins instead.
Read this way, the board tightens and TabFM moves ahead on this column: \mitra{} gives up
7.1\% to the per-task best on average, level with the four-hour AutoGluon
portfolio (7.4\%) and with a smaller gap than EXAONE Tabular (8.3\%), while TabFM's steadier
profile yields the smallest gap (5.3\%). The split boards refine the picture:
on regression \mitra{} has the smallest gap of any single model (1.5\% against
TabFM's 2.4\%; \Cref{tab:reg_headline}), whereas on classification, although
\mitra{} edges TabFM on Elo, TabFM keeps the smaller gap-to-best (6.3\%
against 9.0\%; \Cref{tab:cls_headline}), reflecting its steadier per-task
profile. These differences all lie inside the overlapping bootstrap intervals,
consistent with reading the top of the board as a statistical tie.

\begin{table}[p]
\centering
\small
\caption{Combined TabArena board (51 datasets, 816 units): the top 24 of the
43-method pool, ranked by Elo; lower average
rank is better. The parenthetical marks the configuration setting (default; tuned;
or tuned and ensembled, abbreviated ``tuned + ens.''); the AutoGluon rows are
four-hour AutoML portfolios rather than single foundation models, and the TabFM+
ensemble is excluded from main-text boards per the display policy of
\Cref{sec:setup}. \mitra{} is the default fine-tuned and bagged system. All 24
rows shown have 0\% imputation; the complete 43-method pool, including TabFM+ and
the partially imputed tail, is in \Cref{app:boards}
(\Cref{tab:board_overall_full}). \emph{Gap to best} is the mean relative gap to
the best entry of the pool on each task (TabArena's improvability, in \%): unlike
Elo and average rank, which count wins and losses, it retains the margins, and 0
would mean placing first on every task; lower is better.}
\label{tab:overall}
\begin{tabular}{lrrrr}
\toprule
Method & Elo & 95\% int. & Avg.\ rank & Gap to best (\%) \\
\midrule
AutoGluon 1.6 (noncommercial, 4h) & 1{,}792.9 & $+101/-58$ & 8.38 & 7.4 \\
\textbf{\mitra{} (default)} & \textbf{1{,}774.6} & $+95/-70$ & \textbf{9.02} & \textbf{7.1} \\
TabFM (default) & 1{,}773.5 & $+99/-95$ & 9.06 & 5.3 \\
EXAONE Tabular (default) & 1{,}749.4 & $+71/-55$ & 9.96 & 8.3 \\
AutoGluon 1.6 (extreme, 4h) & 1{,}742.0 & $+86/-55$ & 10.25 & 8.1 \\
AutoGluon 1.5 (4h) & 1{,}652.3 & $+67/-59$ & 14.35 & 9.1 \\
TabPFN-3 (default) & 1{,}637.5 & $+70/-49$ & 15.12 & 10.4 \\
TabPFN-2.6 (default) & 1{,}583.8 & $+60/-42$ & 18.23 & 11.8 \\
RealTabPFN-2.5 (tuned + ens.) & 1{,}568.9 & $+60/-50$ & 19.17 & 11.5 \\
TabICLv2 (default) & 1{,}568.5 & $+61/-54$ & 19.19 & 11.3 \\
AutoGluon 1.4 (4h) & 1{,}479.6 & $+46/-43$ & 25.47 & 13.8 \\
RealMLP (tuned + ens.) & 1{,}477.4 & $+45/-42$ & 25.64 & 13.9 \\
TabDPT (tuned + ens.) & 1{,}437.0 & $+58/-44$ & 28.88 & 14.8 \\
TabDPT-Turbo (default) & 1{,}436.2 & $+50/-44$ & 28.94 & 15.0 \\
TabM (tuned + ens.) & 1{,}422.3 & $+43/-37$ & 30.10 & 15.1 \\
LightGBM (tuned + ens.) & 1{,}404.0 & $+29/-28$ & 31.65 & 16.0 \\
CatBoost (tuned + ens.) & 1{,}394.1 & $+34/-33$ & 32.51 & 15.6 \\
iLTM (tuned + ens.) & 1{,}380.2 & $+41/-37$ & 33.74 & 16.3 \\
ModernNCA (tuned + ens.) & 1{,}365.6 & $+65/-50$ & 35.03 & 16.6 \\
ChimeraBoost (tuned + ens.) & 1{,}358.5 & $+44/-54$ & 35.66 & 16.9 \\
XGBoost (tuned + ens.) & 1{,}352.2 & $+30/-30$ & 36.23 & 16.7 \\
LimiX (default) & 1{,}346.2 & $+70/-58$ & 36.78 & 16.5 \\
TabSwift (default) & 1{,}333.8 & $+57/-49$ & 37.91 & 17.0 \\
xRFM (tuned + ens.) & 1{,}331.8 & $+43/-40$ & 38.09 & 17.2 \\
\bottomrule
\end{tabular}
\end{table}

\subsection{Classification results}
\label{sec:classification}

\paragraph{Headline leaderboard.}
\begin{table}[!t]
\centering
\small
\caption{TabArena classification board (38 datasets, 594 units); top of a
41-method pool (83 configuration entries). Gap to best as defined in
\Cref{tab:overall}. All rows shown have 0\% imputation; the
complete pool is in \Cref{tab:board_cls_full}.}
\label{tab:cls_headline}
\begin{tabular}{lrrrr}
\toprule
Method & Elo & 95\% int. & Avg.\ rank & Gap to best (\%) \\
\midrule
AutoGluon 1.6 (noncommercial, 4h) & 1{,}760.8 & $+84/-54$ & 9.52 & 9.4 \\
\textbf{\mitra{} (default)} & \textbf{1{,}756.3} & $+120/-78$ & \textbf{9.69} & \textbf{9.0} \\
TabFM (default) & 1{,}755.1 & $+125/-111$ & 9.74 & 6.3 \\
EXAONE Tabular (default) & 1{,}750.8 & $+80/-59$ & 9.91 & 10.0 \\
AutoGluon 1.6 (extreme, 4h) & 1{,}711.1 & $+61/-44$ & 11.59 & 10.1 \\
AutoGluon 1.5 (4h) & 1{,}654.2 & $+78/-67$ & 14.39 & 10.6 \\
TabPFN-3 (default) & 1{,}628.7 & $+74/-62$ & 15.80 & 12.9 \\
TabPFN-2.6 (default) & 1{,}577.3 & $+57/-54$ & 18.94 & 14.2 \\
TabICLv2 (default) & 1{,}570.8 & $+70/-64$ & 19.37 & 13.5 \\
\bottomrule
\end{tabular}
\end{table}

On classification, \mitra{} surpasses TabPFN-3 by roughly 128 Elo and delivers the
state-of-the-art level: \mitra{}, TabFM, and EXAONE Tabular span about 6 Elo in
total, far inside every bootstrap interval, so the top of the single-model board
is a three-way statistical tie. The only entry above this group is a four-hour
AutoGluon AutoML portfolio, which is not a single model. This tie is reached under a strictly enforced
one-hour budget that TabFM's own artifacts exceed. The rank-one count again distinguishes portfolio consistency from outright wins
(\Cref{sec:overall}). The AutoGluon portfolio above the trio is outright best
on only 1.5 of the 38 datasets. The foundation models win the datasets
outright: TabFM 7.8, \mitra{} 5.5, and EXAONE Tabular 3.5. The three methods are therefore not interchangeable: each is the outright best on
a different group of datasets, and \mitra{} wins more of them than EXAONE
Tabular.

\paragraph{Per-task win/loss diagnostics.}
Beyond the aggregate Elo, a per-task diagnostic scores each dataset by its mean
metric error over all of its splits on the frozen one-hour board (38 datasets,
594 units). \mitra{} records lower error than TabPFN-3 on 31 of the 38 datasets
(two-sided sign test $p = 0.0001$; \Cref{fig:pertask}), consistent with its
aggregate lead. The wins are material in relative terms: on the median dataset
\mitra{}'s error is $2.1\%$ lower, and in geometric mean TabPFN-3's per-dataset
error is $6.2\%$ higher, while most of the losses are marginal. The same diagnostic is consistent with the three-way tie at the top of the
single-model board: against EXAONE Tabular \mitra{} wins 22 of 38 datasets
($p = 0.42$), indistinguishable from chance, while against TabFM it wins 13 of 38 ($p = 0.07$),
so within the tie TabFM is stronger at the task level while \mitra{} retains
its own set of outright wins (the rank-one counts above). On the
13-dataset regression board the analogous counts favor \mitra{} over both
TabPFN-3 and EXAONE Tabular (10 of 13 each), with TabFM ahead 7 to 6. Raw error
magnitudes mix ROC-AUC error and log-loss and are therefore used only as a
descriptive diagnostic, not pooled as though they shared one scale.

\paragraph{Relationship to Mitra-v1.}
Mitra-v1 is present in the same frozen pools as a TabArena reference entry, so
the version comparison is scored under identical conditions. In this pool
Mitra-v1 rates 1{,}350.2 Elo on the one-hour classification board against
\mitra{}'s 1{,}756.3, and 1{,}265.2 on regression against 1{,}985.6. The
per-task view is equally one-sided. Restricted to the datasets where Mitra-v1
produced real (non-imputed) results, \mitra{} has lower mean metric error on
all 26 of the 26 classification datasets (two-sided sign test $p < 10^{-7}$) and
on 6 of 7 regression datasets. Coverage
improves as much as accuracy: Mitra-v1's entries required imputation on 12 of
the 38 classification datasets and 6 of the 13 regression datasets, whereas
\mitra{} runs every task natively with zero imputation.

\paragraph{Large datasets: a resolved weakness.}
Earlier configurations of \mitra{} were weaker on large tables, because each
bagged child fine-tunes for a fixed number of steps regardless of dataset size
and conditions on a capped in-context support at prediction time. The released
configuration extends to classification the treatment previously applied only on
regression (\Cref{sec:regression}): the fine-tuning support cap is raised to
16{,}384 rows and, after the bagged fit, each bag child predicts with its full
outer training fold as in-context support (``heldout in support'';
\Cref{sec:setup}). Most of the resulting gain comes from the heldout-in-support
rule; the raised cap alone is worth under one Elo (\Cref{app:negative}). On the released board the large-sample deficit is gone: across
the ten classification datasets with at least 10{,}000 training rows \mitra{}'s
mean per-dataset rank in the full pool is 5.3, slightly better than its 6.8 on the
28 smaller datasets, and it is the single best method on several of the largest
tables (for example SDSS17 at 52k training rows and HR-analytics at 13k). It remains marginally behind TabFM on the largest tables (trailing on 6 of
these 10 datasets), so the top-of-board tie of \Cref{sec:classification} holds
within this stratum rather than being an artifact of the smaller datasets. What
remains is a ceiling rather than a deficit: the in-context sweep of
\Cref{sec:efficiency} shows two of the three largest tables still improving at the top of
the support range, so further scaling of in-context support is expected to help
there.

\subsection{Regression results}
\label{sec:regression}

\begin{table}[t]
\centering
\small
\caption{TabArena regression board (13 datasets, 222 units); top of a 40-method
pool (82 configuration entries). Gap to best as defined in \Cref{tab:overall}.
All rows shown have 0\% imputation; the complete
pool is in \Cref{tab:board_reg_full}.}
\label{tab:reg_headline}
\begin{tabular}{lrrrr}
\toprule
Method & Elo & 95\% int. & Avg.\ rank & Gap to best (\%) \\
\midrule
AutoGluon 1.6 (noncommercial, 4h)     & 2{,}078.1 & $+208/-155$ & 4.99  & 1.7 \\
AutoGluon 1.6 (extreme, 4h)           & 2{,}015.5 & $+176/-115$ & 6.28  & 2.2 \\
TabFM (default)                       & 1{,}986.8 & $+170/-96$  & 6.96  & 2.4 \\
\textbf{\mitra{} (default)}           & \textbf{1{,}985.6} & $+223/-134$ & \textbf{6.99}  & \textbf{1.5} \\
EXAONE Tabular (default)              & 1{,}879.5 & $+137/-95$  & 10.07 & 3.5 \\
TabPFN-3 (default)                    & 1{,}798.2 & $+212/-125$ & 13.04 & 3.2 \\
AutoGluon 1.5 (4h)                    & 1{,}773.5 & $+136/-88$  & 14.06 & 4.7 \\
Nori-30M (default)                    & 1{,}756.6 & $+119/-78$  & 14.79 & 4.2 \\
RealTabPFN-2.5 (tuned + ens.)         & 1{,}730.3 & $+133/-92$  & 15.98 & 4.1 \\
TabPFN-2.6 (default)                  & 1{,}729.7 & $+94/-50$   & 16.00 & 4.8 \\
TabDPT (tuned + ens.)                 & 1{,}722.1 & $+156/-87$  & 16.36 & 5.0 \\
TabICLv2 (default)                    & 1{,}682.4 & $+224/-138$ & 18.31 & 4.7 \\
\bottomrule
\end{tabular}
\end{table}

On the full TabArena regression protocol (13 datasets, 222 units;
\Cref{fig:hero}, Regression panel), \mitra{} leads TabPFN-3 by roughly 187 Elo,
rates roughly 106 Elo above EXAONE Tabular on the point estimate, and among
single-configuration foundation models is statistically level with TabFM, roughly one Elo behind on the point
estimate. The regression intervals on this
13-dataset board are wide (half-widths of 95 to 223 Elo across the leading
foundation models), so
these separations are point-estimate orderings rather than interval-clean gaps;
the ordering itself, TabFM and \mitra{} level at the top, then EXAONE Tabular,
then TabPFN-3, mirrors the foundation-model ordering on the overall
board. This is the largest
generational change relative to Mitra-v1.

The entries above \mitra{} are the much larger TabFM and two four-hour AutoGluon
AutoML ensembles. Regression is therefore \mitra{}'s strongest relative result: it
leads TabPFN-3 and, on the point estimate, EXAONE Tabular, whereas classification is a three-way
statistical tie with TabFM and EXAONE Tabular. The rank-one counts show this
directly. No entry in the
82-configuration pool is the single best choice on a regression dataset more often than \mitra{}: 3.6
outright wins of the 13 datasets, ahead of the TabFM+ ensemble (2.5) and
of TabFM (1.5) and the strongest AutoGluon portfolio (1.3). EXAONE
Tabular, rated roughly 106 Elo below \mitra{}, wins only 0.2: its rating comes from
finishing close on most datasets rather than from outright wins. Mitra-v1 trailed every leading regression method by a wide margin;
in this generation, no entry in the pool wins outright more often than \mitra{}.

\paragraph{Sources of the regression gain.}
Three changes account for most of the improvement over Mitra-v1. First, regression
uses the 1{,}000-bin distributional cross-entropy head of \Cref{sec:backbone} rather
than a single-output mean-squared-error head, which gives a calibrated distribution
over the target range instead of a point estimate. Second, at prediction time the
in-context support set is capped at 32{,}768 rows, more than the 20{,}480-row cap
used during fine-tuning; this lets large-sample regression tasks condition on more
context at inference and is the single change most responsible for \mitra{}'s
strength on large regression tables. Third, after the bagged fit each bag child
predicts with its full outer training fold as in-context support (``heldout in
support''; \Cref{sec:setup}), which adds roughly 11 Elo to the released regression
board and is the largest single recent contributor. Feature-budget handling is
always enabled for regression: tables wider than the 256-feature budget are reduced
by a train-only 256-component truncated-SVD projection (a uniform width rule,
applied to every dataset that crosses the threshold; on this suite it fires on one
wide dataset); in a diagnostic on an earlier configuration this feature-budget rule
was worth roughly 17 Elo. With these refinements the released regression board
stands at 1{,}985.6.

\subsection{Deployment-time configuration as a tuning axis}
\label{sec:levers}

The results above use a single frozen checkpoint, yet the same weights can be
deployed at very different operating points. Two prediction-time levers account for a large share of
\mitra{}'s standing. Both are plain environment variables on the released
checkpoint, so a practitioner can reproduce or retune them at zero training cost:
\begin{itemize}[leftmargin=1.6em,topsep=2pt,itemsep=1pt]
\item the \emph{support cap}: the maximum number of training rows the model
  conditions on at prediction time (\Cref{sec:beyondcaps}); it is separate from
  the fine-tuning support cap and changes inference only.
\item \emph{train-only feature selection} (FS): for tables wider than the
  256-feature deployment budget, a train-only reduction to 256 features fit on the
  training split alone: top-256 selection for classification, a 256-component
  truncated-SVD projection for regression.
\end{itemize}
\Cref{tab:levers} quantifies what each lever is worth. Within each block of
the table the weights, the evaluation harness, and the comparison pool are
identical. Rows differ only in the lever settings, so the Elo difference
between adjacent rows is attributable to that lever alone.

\begin{table}[t]
\centering
\small
\caption{What the two deployment-time levers are worth on the frozen \mitra{}
checkpoint. Rows within a block differ only in the two environment-variable
levers; weights, harness, and comparison pool are held fixed, and
$\Delta$\,Elo is the gain over the row above. The regression block is
measured on the full 222-unit protocol; feature selection is always enabled
there, so only the support cap is toggled. The classification block is a
38-dataset single-split diagnostic used for attribution, not the
full-protocol board; its final row lands within bootstrap noise of the
release board of \Cref{sec:classification}.}
\label{tab:levers}
\begin{tabular}{lrcrr}
\toprule
Configuration & Support cap & Train-only FS & Elo & $\Delta$\,Elo \\
\midrule
\multicolumn{5}{l}{\emph{Regression (full 222-unit protocol)}} \\
Cap at the fine-tuning value  & 8{,}192   & on  & 1{,}848.4 & --- \\
Cap raised at prediction time & 32{,}768  & on  & 1{,}935.9 & $+$87.5 \\
\midrule
\multicolumn{5}{l}{\emph{Classification (38-dataset single-split diagnostic)}} \\
Base configuration            & 8{,}192   & off & 1{,}678 & --- \\
$+$ feature selection         & 8{,}192   & on  & 1{,}720 & $+$42 \\
$+$ raised support cap        & 16{,}384  & on  & 1{,}742 & $+$22 \\
\bottomrule
\end{tabular}
\end{table}

\paragraph{Regression: the support cap is decisive.}
The regression block toggles exactly one value: whether the prediction-time
support cap stays at the 8{,}192 rows this ablation used during fine-tuning or is
raised to 32{,}768. On the full 222-unit protocol that single setting moves \mitra{}
from 1{,}848.4 to 1{,}935.9 Elo, a gain of 87.5, and it is what flips
\mitra{} from below EXAONE Tabular (1{,}880) to above it: at the low cap the
checkpoint already leads TabPFN-3 but trails EXAONE. The mechanism is simple:
large-sample tasks condition on $4\times$ as much context at inference
(32{,}768 versus the 8{,}192-row fine-tuning cap of this ablation; the released
configuration fine-tunes with a 20{,}480-row cap and predicts with the same
32{,}768). Two notes reconcile this
ablation with the headline of \Cref{sec:regression}: the ablation endpoint
(1{,}935.9) is within a few Elo of the pre-refinement release board
(1{,}942), and the released board reaches 1{,}985.6 because it additionally
applies the truncated-SVD form of the feature-budget rule on wide tables (the
ablation rows used plain top-256 selection), the raised 20{,}480-row fine-tuning
cap, and the heldout-in-support rule of \Cref{sec:setup}, refinements separate
from the prediction-time cap.
Feature selection itself is a
narrow lever on this suite, active on only the single regression dataset
wider than the 256-feature budget, which is why it stays enabled in both
rows.

\paragraph{Classification: the two levers stack.}
The classification block applies the two levers cumulatively to one checkpoint. Starting from
the base configuration at 1{,}678 Elo, enabling train-only feature selection
adds roughly 42 Elo (to 1{,}720), and raising the support cap from 8{,}192 to
16{,}384 rows adds a further 22 (to 1{,}742), about 64 Elo in total, enough
to move the model into a statistical tie with EXAONE Tabular. Both
levers matter here because the suite contains both wide tables (hundreds to
thousands of features) and large-sample tables. Feature selection is
dtype-gated: it fires on wide continuous tables and is skipped on wide binary
or categorical tables, where reducing to 256 columns would discard signal.
The gate is a uniform rule applied to every dataset, never a per-dataset
choice. Because this block is a single-split diagnostic, its magnitudes are
approximate. Its final row lands within bootstrap noise of the full-protocol
release board of \Cref{sec:classification}, which runs both levers on by
default.

\paragraph{Takeaway and cost.}
The same released weights can be moved by tens of Elo purely through
deployment configuration, and the useful setting depends on dataset shape:
wide tables benefit from feature selection, large-sample tables from a higher
support cap. These settings cost inference throughput, not accuracy: a
larger support set makes each prediction more expensive. Raising the cap is
also a safe default: under GPU memory pressure it halves automatically down to
an 8{,}192-row floor.

\subsection{Accuracy--time trade-off}
\label{sec:efficiency}

\mitra{}'s operating point spends more compute per dataset than forward-pass
foundation models: each unit fine-tunes its eight bagged children for up to 50 steps each and
then predicts with the bag. On the release evaluation artifacts the
end-to-end cost of one evaluation unit (fine-tuning plus bagged prediction; the
harness does not time prediction separately) has a median over datasets of about 8.8 minutes
for classification and 9.6 minutes for regression (TabArena records each
dataset's mean unit time; we quote the median of those means, the same
statistic for every method), and the
stronger accuracy is a different operating point rather than a Pareto-dominance
claim over lighter models such as TabPFN-3. The cost tail, however, is
mostly on the largest datasets, and it is controlled by the same lever that
drives large-sample accuracy in \Cref{sec:levers}: the prediction-time support
cap. This subsection measures both effects of that setting: the accuracy gained
from additional in-context support, and the added inference time.

\begin{figure}[t]
\centering
\includegraphics[width=\linewidth]{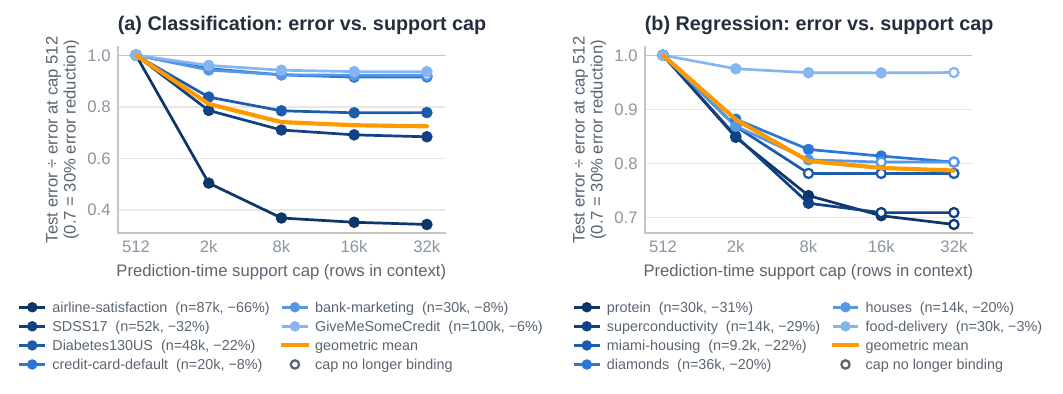}
\caption{In-context sample efficiency of the released checkpoints. Per-dataset
test error (log-loss or $1{-}\mathrm{AUC}$ for classification, RMSE for
regression) as a fraction of its cap-512 value, as the prediction-time support
cap grows: every curve starts at $1.0$, and a value of $0.34$ means the error
has fallen to about a third of its cap-512 level (a $66\%$ reduction). Single
TabArena split, fine-tuning support fixed at 8192, all other settings held
fixed. Legend entries give the training-set size and the total
error reduction. Hollow markers denote caps larger than the rows
available to a bag child ($7/8$ of the training set); from the first hollow
marker onward the curve is exactly flat, an internal control confirming that
only the cap varies.}
\label{fig:nsupport-efficiency}
\end{figure}

\begin{figure}[t]
\centering
\includegraphics[width=\linewidth]{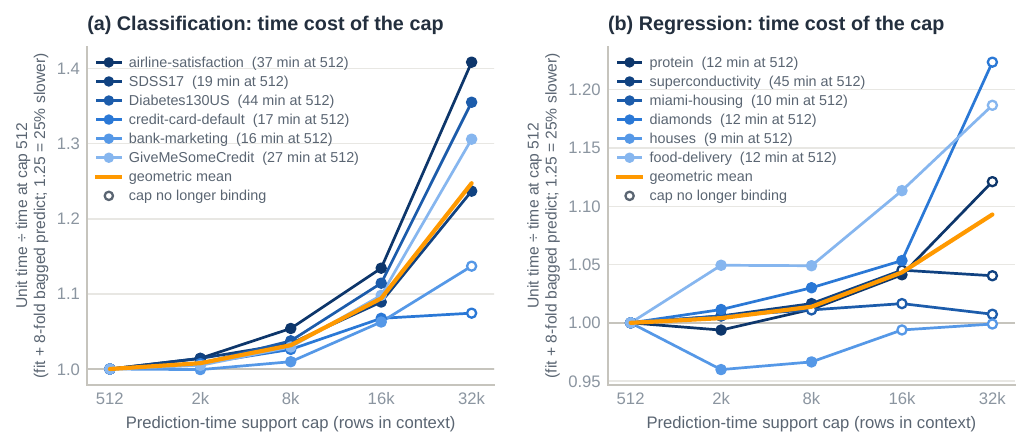}
\caption{Time cost of the same sweep: end-to-end wall-clock of the full
evaluation unit relative to its cap-512 value; legend entries give the
absolute cap-512 time on a single H100. Each unit's cost has two parts that
the curve separates visually: fine-tuning all bagged children, which never
depends on the prediction-time cap and forms the flat floor of every curve,
and bagged prediction, which grows with the cap and accounts for all of the
rise. Hollow markers as in \Cref{fig:nsupport-efficiency}; small excursions
below $1.0$ are run-to-run fit jitter (about $\pm 4\%$) on datasets where the
cap never binds.}
\label{fig:nsupport-cost}
\end{figure}

\paragraph{Setup.}
We sweep the prediction-time support cap over $\{512, 2048, 8192, 16384,
32768\}$ rows on our individual fine-tuned checkpoints, holding fine-tuning and
all other settings fixed. In this sweep fine-tuning sees at most 8192 support
rows, so the sweep varies only how many training rows the fitted model conditions
on when it predicts. Each cell re-runs the complete fit-plus-bagging unit on one TabArena
split (repeat~0, fold~0) of the six largest-sample classification datasets
(20k--100k training rows) and the six largest-sample regression datasets
(9.2k--36k), the tasks where the cap can actually bind. The sweep has a built-in
control: a bag child can see at most $7/8$ of the training rows, so once the cap
exceeds that value it stops binding, and the harness indeed returns
bit-identical errors across all non-binding caps (hollow markers in
\Cref{fig:nsupport-efficiency,fig:nsupport-cost}), confirming that the cap is
the only thing the sweep changes.

\paragraph{In-context sample efficiency.}
\Cref{fig:nsupport-efficiency} shows the accuracy side. Error falls
monotonically as the model is allowed to condition on more training rows (the
only reversals, at the final step, are below $0.2\%$ relative), and the effect
is large: going from cap 512 to 32768 cuts error by $28\%$ for classification
and $21\%$ for regression in geometric mean. The gain is strongly
dataset-dependent. The largest tasks keep improving through the top of the
sweep: airline-satisfaction ($n{=}87\mathrm{k}$) loses $66\%$ of its
$1{-}\mathrm{AUC}$ and SDSS17 ($n{=}52\mathrm{k}$) $32\%$ of its log-loss,
while quickly-saturating tasks such as Food-Delivery-Time flatten by cap 2048
at a $3\%$ total gain. Airline-satisfaction and SDSS17, two of the three largest classification
datasets, are still gaining at 32768 rows. This indicates that the released weights remain support-limited, not
capacity-limited, on big-$n$ tasks: available in-context support, not model capacity,
limits accuracy there. That is why raising the deployment cap (and not further
training) was the cheapest way to obtain the large-sample gains of \Cref{sec:levers}.

\paragraph{What the accuracy costs.}
\Cref{fig:nsupport-cost} shows the price, and its shape follows from the two
cost components of a unit. Fine-tuning does not depend on the prediction-time
cap and dominates the unit, so it forms a flat floor; prediction is the only
part that grows with the cap. The two are also paid on different schedules in
a deployment: the fine-tuning floor is paid once per dataset, while the
prediction increment recurs on every batch of predictions served at the
larger cap. On the sweep the decomposition reads directly off the curve:
through cap 8192 the total is nearly flat (geometric mean $+3\%$
classification, $+1\%$ regression), and the full step to 32768 adds $+25\%$
end-to-end time in geometric mean for classification (worst case $+41\%$ on
airline-satisfaction) and $+9\%$ for regression (worst case $+22\%$ on
diamonds), all of it prediction cost.

\paragraph{Choosing an operating point.}
Putting the two figures together:
cap 8192 already realizes $94\%$ (classification) and $92\%$ (regression) of
the total error reduction at essentially no extra cost, and the release caps
(up to 32{,}768 rows: 32{,}768 on regression and multiclass classification,
16{,}384 on binary classification) obtain most of the remaining accuracy at a
bounded additional time cost. The release therefore ships near the accuracy end
of the curve. A latency-sensitive
deployment can move back down it, e.g.\ to cap 8192 at about a $2\%$
relative-error penalty, by changing one environment variable, with no
retraining. A dedicated low-cost preset that also removes the fine-tuning cost
(zero-shot or reduced-fold operation) is future work, and no result for it is
claimed here.

\begin{figure}[t]
\centering
\includegraphics[width=\linewidth]{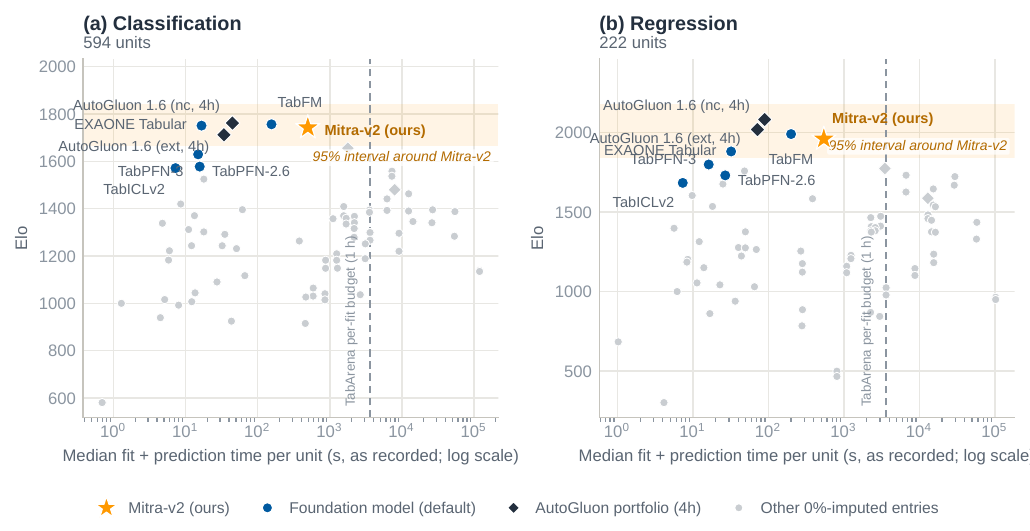}
\caption{Accuracy versus recorded cost on the frozen boards: (a) classification,
(b) regression. Each point is one $0\%$-imputed leaderboard entry; the $x$-axis
is the median over datasets of the mean fit-plus-prediction time per evaluation unit, as recorded in the
reference artifacts (log scale). For \mitra{} the harness does not time
prediction separately, so its recorded time is already the end-to-end unit
cost. Comparator foundation-model defaults are blue circles, AutoGluon
AutoML portfolios are dark diamonds, and \mitra{} is the orange star; the shaded
orange band is the 95\% bootstrap interval around \mitra{}'s rating, and TabFM+ is
excluded as in the main-text boards. The top of each panel is a statistical
cluster inside that band, reached at per-unit costs spanning more than an
order of magnitude; no entry that spends more than \mitra{} reaches the
band, and every member of the band sits well inside the benchmark's one-hour
per-fit budget (dashed vertical line). No Pareto frontier is drawn: recorded
times mix hardware and for TabFM include fits that exceed the one-hour
budget (\Cref{sec:setup}), so the panels compare operating points only.}
\label{fig:elo-time}
\end{figure}

\paragraph{What the cost axis measures.}
\Cref{fig:elo-time} places every $0\%$-imputed entry of the main-text pool in the
accuracy--time plane, using the fit and prediction times recorded in the
frozen reference artifacts. Two properties of the axis matter for reading
the panels. First, it is a \emph{one-time, per-dataset} cost,
paid when a model is fitted to a new table, and it is dominated by fitting
for every method: recorded prediction time is small for every comparator
(median under $5$\,s per unit for the forward-pass foundation models,
$20$--$30$\,s for TabFM, against fit medians of tens to hundreds of
seconds), and \Cref{fig:nsupport-cost} shows the same for \mitra{}, whose
unit cost is dominated by fine-tuning. The recurring cost of serving
predictions is therefore not what separates the methods here. Second,
foundation-model pretraining is excluded for every method, ours included.

\paragraph{Position in the accuracy--time plane.}
The top of both boards is reached at very different costs.
Forward-pass models such as EXAONE Tabular and TabPFN-3 fit in seconds per
unit. \mitra{} spends minutes, almost all of it on fine-tuning and bagging.
The four-hour AutoGluon portfolios and several tuned-and-ensembled baselines
spend up to hours on their slowest tasks. Among these entries,
\mitra{} rates above every method that spends more than it does: every
leaderboard entry with a higher recorded cost also rates below it, and indeed
below the shaded interval that holds the top cluster in \Cref{fig:elo-time}.
This covers all 43 configuration entries with a higher recorded cost on the
classification panel and all 40 on the regression panel, spanning the tuned and ensembled classical field and the
tuned foundation-model configurations.
On this benchmark, no entry that spends more per dataset than \mitra{}'s
roughly nine minutes achieves a higher rating; the fine-tune-and-bag
operating point is the most expensive configuration within the accuracy range it occupies.
The entries that sit to its left at comparable ratings are of two kinds: the
single-forward-pass foundation models that are statistically tied with it
(\Cref{sec:classification}), and the four-hour AutoGluon portfolios, which
are multi-model AutoML systems rather than single models.

\paragraph{Cost inside the band.}
The operating points inside the band differ in one-time cost.
On the same per-dataset median, EXAONE Tabular fits in seconds per unit, TabFM in two to three minutes, and
\mitra{} in about nine to ten at its release defaults. Yet every member of
the band sits well inside the
one-hour per-fit budget that the protocol allows (dashed vertical line), so
within the band, cost is an operational choice rather than a protocol
constraint. Part of \mitra{}'s cost is adjustable: lowering the support cap to
8192 removes most of the added prediction time at about a $2\%$
relative-error cost, and it takes a single runtime setting
(\Cref{sec:efficiency}).

\FloatBarrier
\section{A broader benchmark: TALENT}
\label{sec:talent}

TabArena is our primary benchmark. As an independent check under a
deliberately different setting, we also evaluate \mitra{} on TALENT
\citep{ye2025talent}, a suite of roughly 300 real tabular datasets spanning
classification and regression. Unlike TabArena, TALENT has no time-limit protocol. Every method runs its author-recommended default configuration, and each dataset has one fixed 64/16/20 train/validation/test split, over which we run 15 seeds (0--14) per method.
Because TALENT has no compute budget and no per-dataset tuning, it tests whether \mitra{}'s TabArena results hold up outside the protocol it was developed against. It does: under both matched context policies defined below, \mitra{} sits in the leading group with TabFM and EXAONE Tabular, statistically indistinguishable from both, and on many-class classification it ranks first outright.

\subsection{Protocol and matched-policy design}
\label{sec:talent_protocol}

\paragraph{Presets and slices.}
TALENT's full suite contains 300 datasets, which we score as the
\emph{full300} preset. Our primary preset, \emph{main274}, is the 274 that
remain after excluding the 26 datasets that served as TabPFN-2/TabICLv2
development data (15 classification and 11 regression datasets, matched
against those papers' own disclosures). This follows the same
300\,$-$\,26\,$=$\,274 protocol that the TabPFN-3 and TabICLv2 reports
adopt. The two preset names simply record their dataset counts. We report main274 as the headline
and full300 as a robustness check. The suite splits into three slices that are never mixed:
classification with at most ten classes (173 datasets, scored by accuracy),
classification with more than ten classes (12 datasets, accuracy), and
regression (89 datasets, RMSE). Rankings use the same Elo machinery as
TabArena: a Bradley--Terry fit over per-dataset pairwise comparisons,
anchored at RandomForest~$=$~1{,}000, with 95\% intervals from a
2{,}000-round bootstrap.

\paragraph{Comparison pool.}
The comparison pool is the native TALENT baseline
suite (boosted trees, RealMLP, ModernNCA, TabR, FT-Transformer, and other
neural baselines) plus the tabular foundation models. TabPFN-3 and TabICLv2
are the public models evaluated under the same fixed splits and seeds as
every other method. \mitra{} enters as a single
method: the released classification and regression models in the
default deployment configuration of \Cref{sec:setup}, exactly the
configuration behind the TabArena boards. It is fine-tuned per its
standard recipe with eight-fold bagging, and more-than-ten-class tasks are
handled by the hierarchical decomposition of \Cref{sec:beyondcaps}. Rare
failed cells in the pool are imputed (KNN-5 within slice for
classification, median for regression) and disclosed per method; \mitra{}
has zero imputed cells on main274.

\paragraph{Context composition for in-context models.}
One protocol dimension requires explicit disclosure, because TALENT's fixed
splits make it visible. TabFM and EXAONE Tabular use the combined train
and validation splits as their in-context support set: that is
the documented policy of the drivers that produced their numbers, and it is
those models' recommended usage, since a pure in-context learner has no
other use for a validation split. Classical baselines likewise use the
validation split through early stopping or hyperparameter selection per
their defaults. The TabPFN-3 and TabICLv2 rows, by contrast, use train-only
context: the TALENT integrations of both models fit on the train split
alone (TabPFN-3 uses the validation split only to tune the decision
threshold on binary tasks). A control test that folds the
validation rows into TabPFN-3's support set slightly degraded it, so
train-only is also that model's stronger configuration.

\paragraph{Two matched-policy boards.}
Because context composition materially affects the results, we do not report a single mixed-policy board. Instead we score the
full pool twice, under matched policies. On the \emph{train-only} board,
every in-context model uses the train split alone: \mitra{} runs its
default configuration (the validation split is used only for early
stopping during fine-tuning, and every bag member predicts with the full
train split as its in-context support), and TabFM and EXAONE are re-run
with train-only context over the same 15 seeds. On the \emph{train$+$val}
board, every model that benefits from validation rows in context uses
them: TabFM and EXAONE at their documented defaults, and \mitra{} with the
validation rows folded into every bag member's support set at prediction
time. Fine-tuning and its early stopping are identical to the train-only
run, so the two \mitra{} rows differ only in the rows the model attends to
at prediction time; nothing is fitted or selected on the added rows.
TabPFN-3 and TabICLv2 keep their
train-only configuration on both boards, since that is their stronger one.

\paragraph{The context effect is symmetric across models.}
The re-runs also confirm the context effect is protocol-generic and
roughly symmetric across models: switching to train-only context costs
TabFM 0.20 accuracy points on average over the at-most-ten-class slice
(110 of 173 datasets degrade, 49 improve) and costs EXAONE 0.36 points
(124 degrade, 39 improve), bracketing the 0.29-point gain \mitra{} obtains
from folding the validation rows in (105 of 173 datasets improve, 56
degrade). On the regression slice, train-only context degrades TabFM on 70
of 89 datasets (a 1.8\% geometric-mean RMSE increase) and EXAONE on 75 of
89 (3.2\%), while folding the validation rows in improves \mitra{} on 70
of 89 (a 2.1\% decrease).

\subsection{Results}
\label{sec:talent_results}

\paragraph{Overall standing.}
\Cref{tab:talent_joint} shows the joint main274 board under both context
policies, and the headline is stable across them: \mitra{} sits inside the
leading group's overlapping intervals under both. With train-only context
\mitra{} rates 1{,}479.7 Elo against TabFM's 1{,}503.2 (average rank 4.39
against 4.03), a 24-point gap against bootstrap intervals more than a
hundred points wide, and 28 points above EXAONE Tabular's point estimate (1{,}451.8).
With train$+$val context the top three span 38 points, TabFM 1{,}527.7,
EXAONE Tabular 1{,}519.0, \mitra{} 1{,}490.2, and each point estimate lies
inside the other two models' intervals: the same statistically
indistinguishable group that TabArena produces (\Cref{sec:classification}).
The policy switch mainly moves EXAONE: its rating rises by 67 Elo when
validation rows enter its context (against 25 for TabFM and 11 for
\mitra{}), taking it from below \mitra{} on the train-only board to above
it on the train$+$val board, with the intervals overlapping throughout.
This is consistent with the reverse ablation above, where EXAONE is the
model most sensitive to context composition.

\paragraph{Below the top group.}
Below the top
group the separation is wide on both boards: \mitra{} leads TabPFN-3 by
65--85 Elo and TabICLv2 by 80--96 depending on the policy, and every
foundation model outrates the best classical baseline (ModernNCA,
1{,}166.9) by more than 225 Elo, on a benchmark whose native baselines run
their recommended defaults with no time pressure.

\paragraph{The full suite.}
The unfiltered full300
preset (\Cref{tab:talent_full300}) reproduces the train$+$val picture
(TabFM 1{,}516.6, EXAONE 1{,}510.4, \mitra{} 1{,}488.6, all three within
each other's intervals). Its train-only
re-runs cover the main274 datasets only, so TabFM's imputed-cell count rises
to 38 of 300 there, and we do not lean on that board. One caveat keeps main274 as the headline preset: 26 of the 300
datasets are TabPFN-2/TabICLv2 development datasets, so those two models are
not cleanly comparable on the full suite (\Cref{tab:talent_full300} marks
them). TabFM's 12 imputed cells on the train$+$val boards are the many-class
tasks it cannot run; its train-only main274 board has one more, the binary
eye\_movements\_bin dataset, for which the train-only run produced no result.

\paragraph{Provenance.}
The full suite nonetheless shows a provenance distinction. \mitra{} and
EXAONE Tabular are the only members of the leading group with zero imputed
cells across all 300 datasets. Both also pretrain on entirely synthetic
data, \mitra{} as described in this report and EXAONE Tabular per its
technical report \citep{lgai2026exaonetabular}, so for neither model can an
evaluation dataset have contributed to pretraining. TabPFN-3 and TabICLv2,
also synthetically pretrained, document exactly the 26-dataset development
overlap that main274 removes. TabFM has, at the time of writing, published
no documentation of its training data, so the corresponding guarantee
cannot be checked for it either way.

\begin{table}[t]
\centering
\small
\caption{TALENT joint board, main274 preset (274 datasets: 173
classification with at most ten classes, 12 with more than ten, 89
regression), 15 seeds per method on fixed splits, no time limit, all
methods at author-recommended defaults; top 12 of the pool by Elo
(RandomForest anchored at 1{,}000), scored under the two matched context
policies of the text. On the train-only board every in-context model
consumes the train split alone (TabFM and EXAONE re-run accordingly); on
the train$+$val board TabFM, EXAONE, and \mitra{} fold the validation rows
into context, while TabPFN-3 and TabICLv2 keep their (stronger) train-only
configuration in both. Classical baselines are identical inputs to both
pools; their ratings shift by under 2 Elo between the fits. ``Imp.''
counts imputed dataset cells out of 274 on the train-only\,/\,train$+$val
board.}
\label{tab:talent_joint}
\begin{tabular}{lrrrrrrc}
\toprule
 & \multicolumn{3}{c}{Train-only context} & \multicolumn{3}{c}{Train$+$val context} & \\
\cmidrule(lr){2-4}\cmidrule(lr){5-7}
Method & Elo & 95\% int. & Rank & Elo & 95\% int. & Rank & Imp. \\
\midrule
TabFM                      & 1{,}503.2 & $+62/-56$ & 4.03 & 1{,}527.7 & $+67/-58$ & 3.82 & 13\,/\,12 \\
\textbf{\mitra{}}          & \textbf{1{,}479.7} & $+60/-54$ & \textbf{4.39} & \textbf{1{,}490.2} & $+66/-60$ & \textbf{4.35} & 0\,/\,0 \\
EXAONE Tabular             & 1{,}451.8 & $+52/-46$ & 4.83 & 1{,}519.0 & $+57/-52$ & 3.94 & 1\,/\,0 \\
TabPFN-3                   & 1{,}414.8 & $+54/-50$ & 5.48 & 1{,}404.8 & $+52/-48$ & 5.78 & 1\,/\,1 \\
TabICLv2                   & 1{,}399.6 & $+54/-49$ & 5.76 & 1{,}393.9 & $+53/-49$ & 5.99 & 1\,/\,1 \\
ModernNCA                  & 1{,}166.9 & $+43/-42$ & 11.48 & 1{,}166.7 & $+43/-42$ & 11.54 & 6\,/\,6 \\
RealMLP                    & 1{,}166.1 & $+45/-43$ & 11.50 & 1{,}166.1 & $+45/-43$ & 11.56 & 0\,/\,0 \\
CatBoost                   & 1{,}153.2 & $+30/-29$ & 11.89 & 1{,}154.7 & $+30/-29$ & 11.90 & 0\,/\,0 \\
TabR                       & 1{,}142.6 & $+47/-44$ & 12.22 & 1{,}143.3 & $+48/-43$ & 12.25 & 0\,/\,0 \\
LightGBM                   & 1{,}115.9 & $+29/-29$ & 13.06 & 1{,}116.0 & $+29/-28$ & 13.10 & 0\,/\,0 \\
XGBoost                    & 1{,}091.9 & $+27/-25$ & 13.83 & 1{,}093.2 & $+26/-26$ & 13.83 & 0\,/\,0 \\
FT-Transformer             & 1{,}062.2 & $+40/-41$ & 14.81 & 1{,}062.4 & $+39/-41$ & 14.84 & 0\,/\,0 \\
\bottomrule
\end{tabular}
\end{table}

\begin{table}[p]
\centering
\small
\caption{TALENT main274 per-slice Elo (average rank in parentheses) for the
foundation-model group, under both matched context policies. Slices are
rated independently and are never mixed: accuracy scores the two
classification slices, RMSE the regression slice. TabFM is absent from the
more-than-ten-class slice by construction (its API caps the class count at
ten). TabPFN-3 and TabICLv2 use train-only context in both blocks, their
stronger configuration; classical pool ratings shift by under 3 Elo
between the two fits on the two large slices (under 12 on the 12-dataset
many-class slice).}
\label{tab:talent_slices}
\begin{tabular}{lrrr}
\toprule
 & Cls $\le$10 classes & Cls $>$10 classes & Regression \\
Method & (173 datasets) & (12 datasets) & (89 datasets) \\
\midrule
\multicolumn{4}{l}{\emph{Train-only context}} \\
TabFM                 & 1{,}518.8 (3.93) & ---              & 1{,}480.2 (4.01) \\
\textbf{\mitra{}}     & \textbf{1{,}444.2 (5.16)} & \textbf{2{,}070.1 (2.08)} & \textbf{1{,}499.3 (3.75)} \\
EXAONE Tabular        & 1{,}432.6 (5.38) & 1{,}770.1 (5.25) & 1{,}466.7 (4.20) \\
TabPFN-3              & 1{,}404.1 (5.94) & 1{,}850.3 (4.17) & 1{,}395.3 (5.33) \\
TabICLv2              & 1{,}402.2 (5.98) & 1{,}799.5 (4.83) & 1{,}357.0 (6.01) \\
\midrule
\multicolumn{4}{l}{\emph{Train$+$val context}} \\
TabFM                 & 1{,}529.3 (3.88) & ---              & 1{,}529.8 (3.54) \\
EXAONE Tabular        & 1{,}500.8 (4.30) & 1{,}835.0 (4.58) & 1{,}532.6 (3.51) \\
\textbf{\mitra{}}     & \textbf{1{,}448.7 (5.18)} & \textbf{2{,}204.2 (1.58)} & \textbf{1{,}516.1 (3.71)} \\
TabPFN-3              & 1{,}396.0 (6.20) & 1{,}841.7 (4.50) & 1{,}382.0 (5.71) \\
TabICLv2              & 1{,}395.5 (6.21) & 1{,}809.2 (4.92) & 1{,}352.5 (6.24) \\
\bottomrule
\end{tabular}
\end{table}

\begin{table}[p]
\centering
\small
\caption{TALENT joint board, \emph{full300} preset (all 300 datasets, no
exclusions), scored under the train$+$val context policy (the train-only
re-runs of TabFM and EXAONE cover the main274 datasets only, so a
train-only full300 board would impute their 26 excluded cells; the text
mentions it only to note its imputation count); same protocol, pool,
and columns as \Cref{tab:talent_joint}, top 12 by Elo, with ``Imp.''
counted out of 300. Rows marked $^{*}$ are not cleanly comparable on this
preset (see note below).}
\label{tab:talent_full300}
\begin{tabular}{lrrrrr}
\toprule
Method & Elo & 95\% int. & Avg.\ rank & Win rate & Imp. \\
\midrule
TabFM                      & 1{,}516.6 & $+64/-58$ & 3.86  & 0.905 & 12 \\
EXAONE Tabular             & 1{,}510.4 & $+55/-51$ & 3.94  & 0.902 & 0  \\
\textbf{\mitra{}}          & \textbf{1{,}488.6} & $+69/-59$ & \textbf{4.26} & \textbf{0.891} & \textbf{0} \\
TabPFN-3$^{*}$             & 1{,}369.8 & $+50/-45$ & 6.34  & 0.822 & 26 \\
TabICLv2$^{*}$             & 1{,}361.8 & $+49/-45$ & 6.50  & 0.817 & 26 \\
RealMLP                    & 1{,}167.6 & $+41/-39$ & 11.43 & 0.652 & 0  \\
ModernNCA                  & 1{,}164.0 & $+39/-39$ & 11.54 & 0.649 & 6  \\
CatBoost                   & 1{,}155.4 & $+28/-27$ & 11.80 & 0.640 & 0  \\
TabR                       & 1{,}150.1 & $+44/-40$ & 11.96 & 0.635 & 0  \\
LightGBM                   & 1{,}116.2 & $+28/-28$ & 13.02 & 0.599 & 0  \\
XGBoost                    & 1{,}091.8 & $+24/-25$ & 13.80 & 0.573 & 0  \\
FT-Transformer             & 1{,}065.7 & $+38/-37$ & 14.66 & 0.545 & 0  \\
\bottomrule
\end{tabular}

\vspace{3pt}
{\footnotesize\raggedright $^{*}$\,26 of the 300 datasets are
TabPFN-2/TabICLv2 development sets, the exact set that \emph{main274} removes,
following the 300\,$-$\,26\,$=$\,274 protocol that those models' own reports
adopted (documented in Appendix~E.3 of the TabPFN-3 report
\citep{grinsztajn2026tabpfn3} and Appendix~K of the TabICLv2 paper
\citep{qu2026tabiclv2}). We do not run those two models on their own
development data, so their 26 cells are imputed; their full300 rating is therefore not a
like-for-like comparison, which is why main274 is our headline preset.
\mitra{}, trained only on synthetic priors, overlaps no dataset and imputes no
cell.\par}
\end{table}

\paragraph{Results by slice.}
\Cref{tab:talent_slices} decomposes both boards. The more-than-ten-class
slice is \mitra{}'s strongest result on the suite: it ranks \emph{first in
the entire pool}, ahead of every foundation model and every classical
baseline, under both context policies, at 2{,}070.1 Elo (average rank
2.08) with train-only context and 2{,}204.2 (rank 1.58) with train$+$val,
leading the runner-up (TabPFN-3) by 220 and 363 Elo respectively. It does
so despite never seeing a task with more than ten classes during
pretraining: the balanced hierarchical decomposition of
\Cref{sec:beyondcaps} extends the frozen ten-class head to these targets
at deployment time, with no additional training. TabFM cannot run the
slice at all. The intervals on this 12-dataset slice are wide, so we read
the margins as favorable point estimates rather than a decisive
separation; the rank-one position itself, however, holds under both
policies. On regression, \mitra{} posts the highest point estimate in
the pool on the train-only board (1{,}499.3, against TabFM's 1{,}480.2 and
EXAONE's 1{,}466.7), and the three form a statistical tie with train$+$val
context (EXAONE 1{,}532.6, TabFM 1{,}529.8, \mitra{} 1{,}516.1), in both
cases far ahead of TabPFN-3 and TabICLv2.

\paragraph{The at-most-ten-class slice.}
This slice is
\mitra{}'s weakest on either board, and the source of its joint-board gap
to TabFM: it trails TabFM by 75 Elo with train-only context (1{,}444.2
against 1{,}518.8) and by 81 with train$+$val (1{,}448.7 against
1{,}529.3), with mean accuracy gaps of 0.12 and 0.19 points
respectively, consistent gaps that remain within per-slice interval
widths of roughly $\pm$70--80. Reading across policies would overstate
this deficit: a mixed-policy comparison of \mitra{} train-only against
TabFM's train$+$val default scores it at roughly 85 Elo, about 10 points
of which is the context-policy difference rather than modeling. \mitra{}
places second or third in this slice's foundation-model group depending
on policy, above EXAONE without validation context (1{,}444.2 against
1{,}432.6) and below it with (1{,}448.7 against 1{,}500.8), the same
EXAONE swing that drives the joint-board reordering.

\FloatBarrier
\section{Limitations}
\label{sec:limitations}

\begin{description}[leftmargin=0pt,style=nextline]
\item[Overlapping uncertainty on an evolving benchmark.] The bootstrap intervals
across the leading group overlap. In particular, the gaps separating \mitra{},
TabFM, and EXAONE Tabular on the classification and overall boards are well
inside those intervals, so the top of each board is best read as a statistical
tie rather than a strict ordering, and the ``state-of-the-art level'' framing
used in this report should be read the same way. The regression board rests on
13 datasets and has the widest intervals, so its separations, including the
lead over EXAONE Tabular, are point-estimate orderings. TabArena itself evolves
as methods, datasets, and scorer logic change: the frozen revision is necessary
for reproduction, and future public rankings may differ.

\item[Downstream compute.] The headline system uses fine-tuning and eight-fold
bagging. It is much slower than forward-pass foundation models and should not
be described as zero-shot or Pareto-dominant (\Cref{fig:elo-time}).

\item[Incomplete training provenance.] \mitra{} changes context length, feature
range, prior mixture, optimizer, distributed recipe, and checkpoint at once,
and the current experiments do not isolate the contribution of each change.
Accelerator hours, energy, and the realized prior mixture frequencies are
likewise not recoverable from the current metadata.

\item[Large-dataset headroom.] Earlier configurations of \mitra{} were weaker on
the datasets with tens of thousands of training rows, because each bagged child
fine-tunes for a fixed number of steps regardless of dataset size. The released
configuration closes this deficit mainly through the heldout-in-support rule,
alongside a raised classification fine-tuning support cap (\Cref{sec:classification}),
so on the released board \mitra{}'s mean per-dataset rank on the large-sample
classification datasets is slightly better than on the smaller ones. What remains
is a ceiling rather than a deficit: the in-context sweep of \Cref{sec:efficiency}
shows two of the three largest tables still improving at the top of the support range, so
those tables retain headroom that further scaling of in-context support may
recover.

\item[Timing instrumentation and evaluation currency.] Prediction time is not
instrumented separately in the release artifacts. As a result, the accuracy--time
analysis of \Cref{sec:efficiency} reports controlled end-to-end wall-clock
(fine-tuning plus bagged prediction) on a single harness, not a separated
prediction-latency figure. Separately, the TALENT results (\Cref{sec:talent})
are scored against a frozen snapshot of that suite and its comparison pool. As
with the TabArena boards, the frozen revision is what makes the numbers
reproducible, and future public rankings may differ as the benchmark and its
baselines evolve.
\end{description}
\section{Conclusion}
\label{sec:conclusion}

\mitra{} is a substantial update built around synthetic-data and optimization
scaling rather than a deeper Transformer. A 12-layer model trained with larger
support/query contexts, more features, and a Hybrid SCM prior surpasses
TabPFN-3 on both task types and matches the leading tabular foundation models
on the full multi-split TabArena protocol: its overall
rating sits at the top of the single-model field, statistically level with TabFM
and EXAONE Tabular while holding the highest point estimate of the three. On classification the three models are in a
statistical tie, and \mitra{} is clearly ahead of TabPFN-3, covering all tasks
without imputation. Regression is the largest improvement over Mitra-v1, which trailed the field by a wide margin: \mitra{} now leads TabPFN-3 by a wide margin, rates above EXAONE Tabular on the point estimate, and is statistically level with TabFM among single-configuration foundation models, under a strictly enforced one-hour budget that TabFM's own leaderboard artifacts exceed. On regression no entry in the full comparison
pool wins a dataset outright more often than \mitra{} (3.6 outright wins of 13, against 2.5 for the four-hour TabFM+ ensemble).
These results also hold off its home benchmark: on TALENT (roughly 300 datasets, fixed splits, no time limit), \mitra{} is statistically indistinguishable from TabFM and EXAONE Tabular at the top under both matched context policies, and on many-class classification it places first in the entire pool, a regime absent from its pretraining and one that TabFM cannot handle at all.

Its limitations are set out in \Cref{sec:limitations}. The system is fine-tuned and bagged rather than zero-shot, its standing sits inside overlapping intervals, and its runtime is well above forward-pass models. These point to the next steps: documenting the full training provenance, recording prediction time separately in the release harness, and pushing in-context support further on the very largest tables, where the sweep of \Cref{sec:efficiency} shows accuracy still rising. The main finding concerns model design: scaling the synthetic task distribution and the optimizer, with the 12-layer backbone left unchanged, delivers the state-of-the-art level of the leading tabular foundation models, at 77M parameters against the 1.6B of TabFM. Neither depth nor parameter count is the limiting factor.

\bibliographystyle{plainnat}
\bibliography{references}

\begin{thebibliography}{22}
\providecommand{\natexlab}[1]{#1}
\providecommand{\url}[1]{\texttt{#1}}
\expandafter\ifx\csname urlstyle\endcsname\relax
  \providecommand{\doi}[1]{doi: #1}\else
  \providecommand{\doi}{doi: \begingroup \urlstyle{rm}\Url}\fi

\bibitem[Arazi et~al.(2025)Arazi, Shapira, and Reichart]{arazi2025tabstar}
Alan Arazi, Eilam Shapira, and Roi Reichart.
\newblock {TabSTAR}: A tabular foundation model for tabular data with text
  fields.
\newblock In \emph{Advances in Neural Information Processing Systems
  (NeurIPS)}, 2025.
\newblock arXiv:2505.18125.

\bibitem[Bouadi et~al.(2026)Bouadi, Bouarour, Kulkarni, Dubey, Tanna, and
  Sankarapu]{bouadi2026shaping}
Mohamed Bouadi, Nassim Bouarour, Varun Kulkarni, Shivam Dubey, Aditya Tanna,
  and Vinay~Kumar Sankarapu.
\newblock Shaping the prior: How synthetic task distributions determine tabular
  foundation model quality.
\newblock \emph{arXiv preprint arXiv:2605.18971}, 2026.

\bibitem[Dao(2026)]{dao2026gramns}
Tri Dao.
\newblock Gram {Newton-Schulz}: A fast, hardware-aware {Newton-Schulz}
  algorithm for {Muon}.
\newblock \url{https://tridao.me/blog/2026/gram-newton-schulz/}, 2026.
\newblock Blog post.

\bibitem[Eo et~al.(2026)Eo, Suh, Cho, Kim, Kim, Nam, and
  Lee]{lgai2026exaonetabular}
Moonjung Eo, Min-Kook Suh, Hye-Seung Cho, Jiwon Kim, Seoyoon Kim, Sangjun Nam,
  and Soonyoung Lee.
\newblock {EXAONE Tabular 1.0}: Technical report.
\newblock \url{https://github.com/LGAI-Research/EXAONE-Tabular}, 2026.
\newblock arXiv:2608.25774.

\bibitem[Erickson et~al.(2020)Erickson, Mueller, Shirkov, Zhang, Larroy, Li,
  and Smola]{erickson2020autogluon}
Nick Erickson, Jonas Mueller, Alexander Shirkov, Hang Zhang, Pedro Larroy,
  Mu~Li, and Alexander Smola.
\newblock {AutoGluon-Tabular}: Robust and accurate {AutoML} for structured
  data, 2020.
\newblock arXiv:2003.06505.

\bibitem[Erickson et~al.(2025)Erickson, Purucker, Tschalzev, Holzm{\"u}ller,
  Desai, Salinas, and Hutter]{erickson2025tabarena}
Nick Erickson, Lennart Purucker, Andrej Tschalzev, David Holzm{\"u}ller,
  Prateek~Mutalik Desai, David Salinas, and Frank Hutter.
\newblock {TabArena}: A living benchmark for machine learning on tabular data,
  2025.
\newblock arXiv:2506.16791.

\bibitem[Fang and Wang(2026)]{exts}
Haoyang Fang and Bernie Wang.
\newblock Exploit more, explore smarter for budget-constrained agentic search,
  2026.
\newblock URL \url{https://arxiv.org/abs/2608.23848}.

\bibitem[Fang et~al.(2026)Fang, Zhu, Han, Zhang, Pan, Yang, Zhang, Gai, Tang,
  Hu, et~al.]{llmzero}
Haoyang Fang, Wei Zhu, Boran Han, Alex Zhang, Zhenyu Pan, Shuo Yang, Shuai
  Zhang, Jiading Gai, Peng Tang, Cuixiong Hu, et~al.
\newblock {LLMZero}: Discovering adaptive training strategies for {RL}
  post-training via {LLM} agents.
\newblock \emph{arXiv preprint arXiv:2606.18388}, 2026.

\bibitem[{Google Research}(2026)]{google2026tabfm}
{Google Research}.
\newblock {TabFM}: A zero-shot foundation model for tabular data.
\newblock \url{https://github.com/google-research/tabfm}, 2026.
\newblock Zero-shot tabular foundation model; blog:
  \url{https://research.google/blog/introducing-tabfm-a-zero-shot-foundation-model-for-tabular-data/}.

\bibitem[Gorishniy et~al.(2025)Gorishniy, Kotelnikov, and
  Babenko]{gorishniy2025tabm}
Yury Gorishniy, Akim Kotelnikov, and Artem Babenko.
\newblock {TabM}: Advancing tabular deep learning with parameter-efficient
  ensembling.
\newblock In \emph{International Conference on Learning Representations}, 2025.

\bibitem[Grinsztajn et~al.(2025)Grinsztajn, Fl{\"o}ge, Key, Birkel, Jund, Roof,
  et~al.]{grinsztajn2025tabpfn25}
L{\'e}o Grinsztajn, Klemens Fl{\"o}ge, Oscar Key, Felix Birkel, Philipp Jund,
  Brendan Roof, et~al.
\newblock {TabPFN-2.5}: Advancing the state of the art in tabular foundation
  models, 2025.
\newblock arXiv:2511.08667.

\bibitem[Grinsztajn et~al.(2026)Grinsztajn, Fl{\"o}ge, Key, Birkel, Jund, Roof,
  et~al.]{grinsztajn2026tabpfn3}
L{\'e}o Grinsztajn, Klemens Fl{\"o}ge, Oscar Key, Felix Birkel, Philipp Jund,
  Brendan Roof, et~al.
\newblock {TabPFN-3}: Technical report, 2026.
\newblock arXiv:2605.13986.

\bibitem[Hollmann et~al.(2022)Hollmann, M{\"u}ller, Eggensperger, and
  Hutter]{hollmann2022tabpfn}
Noah Hollmann, Samuel M{\"u}ller, Katharina Eggensperger, and Frank Hutter.
\newblock {TabPFN}: A transformer that solves small tabular classification
  problems in a second, 2022.
\newblock arXiv:2207.01848.

\bibitem[Hollmann et~al.(2025)Hollmann, M{\"u}ller, Purucker, Krishnakumar,
  K{\"o}rfer, Hoo, Schirrmeister, and Hutter]{hollmann2025tabpfn}
Noah Hollmann, Samuel M{\"u}ller, Lennart Purucker, Arjun Krishnakumar, Max
  K{\"o}rfer, Shi~Bin Hoo, Robin~Tibor Schirrmeister, and Frank Hutter.
\newblock Accurate predictions on small data with a tabular foundation model.
\newblock \emph{Nature}, 637:\penalty0 319--326, 2025.
\newblock \doi{10.1038/s41586-024-08328-6}.

\bibitem[Holzm{\"u}ller et~al.(2024)Holzm{\"u}ller, Grinsztajn, and
  Steinwart]{holzmueller2024realmlp}
David Holzm{\"u}ller, L{\'e}o Grinsztajn, and Ingo Steinwart.
\newblock Better by default: Strong pre-tuned {MLPs} and boosted trees on
  tabular data.
\newblock In \emph{Advances in Neural Information Processing Systems}, 2024.

\bibitem[Peroni et~al.(2025)Peroni, Le, and Sheinin]{peroni2025robust}
Matthew Peroni, Franck Le, and Vadim Sheinin.
\newblock Robust tabular foundation models, 2025.
\newblock arXiv:2512.03307.

\bibitem[Qu et~al.(2025)Qu, Holzm{\"u}ller, Varoquaux, and
  Morvan]{qu2025tabicl}
Jingang Qu, David Holzm{\"u}ller, Ga{\"e}l Varoquaux, and Marine~Le Morvan.
\newblock {TabICL}: A tabular foundation model for in-context learning on large
  data.
\newblock In \emph{International Conference on Machine Learning (ICML)}, 2025.
\newblock arXiv:2502.05564.

\bibitem[Qu et~al.(2026)Qu, Holzm{\"u}ller, Varoquaux, and
  Morvan]{qu2026tabiclv2}
Jingang Qu, David Holzm{\"u}ller, Ga{\"e}l Varoquaux, and Marine~Le Morvan.
\newblock {TabICLv2}: A better, faster, scalable, and open tabular foundation
  model, 2026.
\newblock arXiv:2602.11139.

\bibitem[Spinaci et~al.(2025)Spinaci, Polewczyk, Schambach, and
  Thelin]{spinaci2025contexttab}
Marco Spinaci, Marek Polewczyk, Maximilian Schambach, and Sam Thelin.
\newblock {ConTextTab}: A semantics-aware tabular in-context learner.
\newblock In \emph{Advances in Neural Information Processing Systems
  (NeurIPS)}, 2025.
\newblock Spotlight. arXiv:2506.10707.

\bibitem[Ye et~al.(2025)Ye, Liu, Cai, Zhou, and Zhan]{ye2025talent}
Han-Jia Ye, Si-Yang Liu, Hao-Run Cai, Qi-Le Zhou, and De-Chuan Zhan.
\newblock A closer look at deep learning methods on tabular datasets ({TALENT}:
  A tabular analytics and learning toolbox).
\newblock \emph{Journal of Machine Learning Research}, 26, 2025.
\newblock \url{https://github.com/LAMDA-Tabular/TALENT}.

\bibitem[Zhang et~al.(2025{\natexlab{a}})Zhang, Ren, Yu, Yuan, Wang, Li, Wu,
  Mo, Mao, Hao, Dai, Xu, Li, Zhang, He, Wang, Zhang, Xu, Li, Gao, Zou, Liu,
  Liu, Xu, Cheng, Li, Zhou, Li, Fan, Lin, Han, Li, Lu, Xue, Jiang, Wang, Wang,
  and Cui]{zhang2025limix}
Xingxuan Zhang, Gang Ren, Han Yu, Hao Yuan, Hui Wang, Jiansheng Li, Jiayun Wu,
  Lang Mo, Li~Mao, Mingchao Hao, Ningbo Dai, Renzhe Xu, Shuyang Li, Tianyang
  Zhang, Yue He, Yuanrui Wang, Yunjia Zhang, Zijing Xu, Dongzhe Li, Fang Gao,
  Hao Zou, Jiandong Liu, Jiashuo Liu, Jiawei Xu, Kaijie Cheng, Kehan Li, Linjun
  Zhou, Qing Li, Shaohua Fan, Xiaoyu Lin, Xinyan Han, Xuanyue Li, Yan Lu, Yuan
  Xue, Yuanyuan Jiang, Zimu Wang, Zhenlei Wang, and Peng Cui.
\newblock {LimiX}: Unleashing structured-data modeling capability for
  generalist intelligence, 2025{\natexlab{a}}.
\newblock arXiv:2509.03505.

\bibitem[Zhang et~al.(2025{\natexlab{b}})Zhang, Maddix, Yin, Erickson, Ansari,
  Han, Zhang, Akoglu, Faloutsos, Mahoney, Hu, Rangwala, Karypis, and
  Wang]{zhang2025mitra}
Xiyuan Zhang, Danielle~C. Maddix, Junming Yin, Nick Erickson, Abdul~Fatir
  Ansari, Boran Han, Shuai Zhang, Leman Akoglu, Christos Faloutsos, Michael~W.
  Mahoney, Cuixiong Hu, Huzefa Rangwala, George Karypis, and Bernie Wang.
\newblock {Mitra}: Mixed synthetic priors for enhancing tabular foundation
  models, 2025{\natexlab{b}}.
\newblock arXiv:2510.21204.

\end{thebibliography}

\appendix
\addtocontents{toc}{\protect\setcounter{tocdepth}{1}}
\section{Reproducibility and release artifacts}
\label{app:repro}

\subsection{Frozen evidence}
All numbers in this report are pinned to a frozen evaluation of the released
classifier and regressor weights. The TabArena boards are the one-hour-protocol
boards of 3 September 2026: classification over all 594 TabArena
repeat$\times$fold units and regression over all 222 units, run with a
3{,}600\,s per-unit time limit and a 250\,s per-bag-child fine-tuning budget
(\Cref{sec:setup}), then scored against the full comparator pool with 200
bootstrap rounds and random-forest default fill. The complete leaderboards are
given in \Cref{app:boards}; the TALENT boards are given in \Cref{sec:talent}.

\subsection{Release contents}
The public release consists of three parts, all under the Apache-2.0 license.
\begin{enumerate}[leftmargin=*,itemsep=2pt]
\item \textbf{Model weights.} The classifier and regressor weights, distributed at
  \url{https://huggingface.co/autogluon/mitra-classifier-2} and
  \url{https://huggingface.co/autogluon/mitra-regressor-2}.
\item \textbf{Inference and fine-tuning code.} Distributed at\\
  \url{https://huggingface.co/autogluon/mitra-finetune}. It implements the
  default deployment of \Cref{sec:setup} (short fine-tuning, eight-fold bagging,
  heldout-in-support, feature selection, and the many-class wrapper) with the
  deployment configuration behind every board in this report as its defaults,
  together with a minimal fit/predict example.
\item \textbf{Evaluation results.} The per-unit TabArena results behind the
  boards of \Cref{sec:experiments} and \Cref{app:boards}, published in the
  \texttt{results/} directory of the code repository in the same per-unit format
  that TabFM uses for its released results (one row per evaluation unit, that is,
  per dataset, repeat, and fold, with the metric error, the metric, and the problem
  type; 594 classification rows and 222 regression rows).
\end{enumerate}

The pretraining pipeline (the multi-node trainer, the synthetic-data generators,
and their configurations) is not part of this release. \Cref{sec:model}
describes its design.
\section{Complete TabArena leaderboards}
\label{app:boards}

\begin{figure}[!h]
\centering
\includegraphics[width=0.55\textwidth]{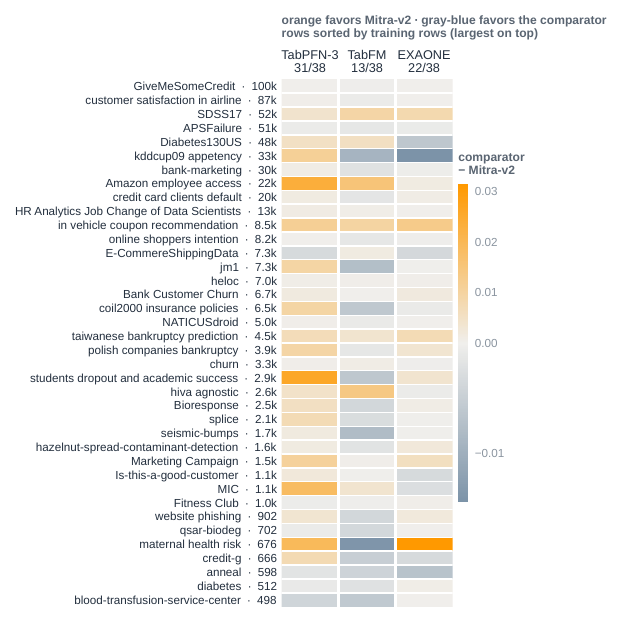}
\caption{Per-task mean metric error difference (comparator $-$ \mitra{})
against TabPFN-3, TabFM, and EXAONE Tabular on the frozen one-hour
classification board; orange favors \mitra{}, and each column header gives
the number of datasets where \mitra{} has lower error. Rows are sorted by
training-set size (largest on top; row labels give training rows).
Magnitudes mix ROC-AUC error and log-loss (descriptive only).}
\label{fig:pertask}
\end{figure}

This appendix gives the full comparison pools behind the figures and headline tables
of \Cref{sec:experiments}. Each table lists one row per method at its best-scoring
configuration, ranked by Elo within the frozen reference pool for that board (overall
covers 51 datasets over 816 units, classification 38 datasets over 594 units, and
regression 13 datasets over 222 units). The pools contain 85, 83, and 82
configuration entries respectively, which collapse to 43, 41, and 40 distinct
methods. The \emph{Regime} column records the configuration protocol: ``Default''
and ``Tuned'' are single models, ``Tuned + ens.'' is a tuned-and-ensembled portfolio,
``AutoML'' marks the AutoGluon four-hour AutoML systems, and ``Ensemble'' marks the
TabFM+ foundation-model ensemble. The \emph{Imp.} column is the fraction of
evaluation units imputed rather than run natively; a dash means full native coverage
(0\% imputed). The \emph{Gap (\%)} column is the mean relative gap to the best
entry of the pool on each task (TabArena's improvability). Unlike Elo, rank, and
win rate, which count wins and losses, it keeps the margins. Lower is better. Intervals are asymmetric 95\% bootstrap intervals from 200 rounds.
Rows with nonzero imputation did not run on every unit and should be read with that
caveat. The 0\%-imputed leaders of each pool are the main-text tables
(\Cref{tab:overall}, \Cref{tab:cls_headline}, \Cref{tab:reg_headline}). \mitra{} is shown
in bold. \Cref{fig:pertask} complements the classification board with the
per-task diagnostic discussed in \Cref{sec:classification}.

\setlength{\LTcapwidth}{\linewidth}

\clearpage
\footnotesize
\setlength{\tabcolsep}{4.5pt}
\renewcommand{\arraystretch}{0.93}
\begin{longtable}{llrrrrrr}
\caption{Complete overall TabArena board (51 datasets, 816 units): all 43 methods,
ranked by Elo.}
\label{tab:board_overall_full}\\
\toprule
Method & Regime & Elo & 95\% int. & Avg.\ rank & Win rate & Gap (\%) & Imp. \\
\midrule
\endfirsthead
\multicolumn{8}{l}{\itshape Overall board (continued from previous page)}\\
\toprule
Method & Regime & Elo & 95\% int. & Avg.\ rank & Win rate & Gap (\%) & Imp. \\
\midrule
\endhead
\midrule
\multicolumn{8}{r}{\itshape continued on next page}\\
\endfoot
\bottomrule
\endlastfoot
TabFM+ & Ensemble & 1{,}817.4 & $+97/-78$ & 7.58 & 0.922 & 5.0 & -- \\
AutoGluon 1.6 (noncommercial, 4h) & AutoML & 1{,}792.9 & $+101/-58$ & 8.38 & 0.912 & 7.4 & -- \\
\textbf{\mitra{}} & Default & 1{,}774.6 & $+95/-70$ & 9.02 & 0.905 & 7.1 & -- \\
TabFM & Default & 1{,}773.5 & $+99/-95$ & 9.06 & 0.904 & 5.3 & -- \\
EXAONE Tabular & Default & 1{,}749.4 & $+71/-55$ & 9.96 & 0.893 & 8.3 & -- \\
AutoGluon 1.6 (extreme, 4h) & AutoML & 1{,}742.0 & $+86/-55$ & 10.25 & 0.890 & 8.1 & -- \\
AutoGluon 1.5 (4h) & AutoML & 1{,}652.3 & $+67/-59$ & 14.35 & 0.841 & 9.1 & -- \\
TabPFN-3 & Default & 1{,}637.5 & $+70/-49$ & 15.12 & 0.832 & 10.4 & -- \\
TabPFN-2.6 & Default & 1{,}583.8 & $+60/-42$ & 18.23 & 0.795 & 11.8 & -- \\
RealTabPFN-2.5 & Tuned + ens. & 1{,}568.9 & $+60/-50$ & 19.17 & 0.784 & 11.5 & -- \\
TabICLv2 & Default & 1{,}568.5 & $+61/-54$ & 19.19 & 0.783 & 11.3 & -- \\
AutoGluon 1.4 (4h) & AutoML & 1{,}479.6 & $+46/-43$ & 25.47 & 0.709 & 13.8 & -- \\
RealMLP & Tuned + ens. & 1{,}477.4 & $+45/-42$ & 25.64 & 0.707 & 13.9 & -- \\
TabDPT & Tuned + ens. & 1{,}437.0 & $+58/-44$ & 28.88 & 0.668 & 14.8 & -- \\
TabDPT-Turbo & Default & 1{,}436.2 & $+50/-44$ & 28.94 & 0.667 & 15.0 & -- \\
TabM & Tuned + ens. & 1{,}422.3 & $+43/-37$ & 30.10 & 0.654 & 15.1 & -- \\
LightGBM & Tuned + ens. & 1{,}404.0 & $+29/-28$ & 31.65 & 0.635 & 16.0 & -- \\
CatBoost & Tuned + ens. & 1{,}394.1 & $+34/-33$ & 32.51 & 0.625 & 15.6 & -- \\
iLTM & Tuned + ens. & 1{,}380.2 & $+41/-37$ & 33.74 & 0.610 & 16.3 & -- \\
ModernNCA & Tuned + ens. & 1{,}365.6 & $+65/-50$ & 35.03 & 0.595 & 16.6 & -- \\
ChimeraBoost & Tuned + ens. & 1{,}358.5 & $+44/-54$ & 35.66 & 0.587 & 16.9 & -- \\
XGBoost & Tuned + ens. & 1{,}352.2 & $+30/-30$ & 36.23 & 0.581 & 16.7 & -- \\
LimiX & Default & 1{,}346.2 & $+70/-58$ & 36.78 & 0.574 & 16.5 & -- \\
TabSwift & Default & 1{,}333.8 & $+57/-49$ & 37.91 & 0.561 & 17.0 & -- \\
xRFM & Tuned + ens. & 1{,}331.8 & $+43/-40$ & 38.09 & 0.558 & 17.2 & -- \\
TabPFNv2 & Tuned + ens. & 1{,}325.5 & $+62/-65$ & 38.67 & 0.552 & 17.6 & 35\% \\
Mitra-v1 & Default & 1{,}314.1 & $+62/-63$ & 39.72 & 0.539 & 18.1 & 35\% \\
TabICL & Default & 1{,}306.9 & $+50/-58$ & 40.39 & 0.531 & 18.0 & 29\% \\
BetaTabPFN & Default & 1{,}270.5 & $+57/-55$ & 43.77 & 0.491 & 19.4 & 25\% \\
SAP-RPT-OSS & Default & 1{,}270.0 & $+56/-56$ & 43.82 & 0.490 & 19.2 & -- \\
NeuralNet (PyTorch) & Tuned + ens. & 1{,}268.9 & $+46/-45$ & 43.93 & 0.489 & 18.1 & -- \\
Explainable Boosting & Tuned + ens. & 1{,}252.0 & $+36/-37$ & 45.50 & 0.470 & 19.6 & -- \\
ExtraTrees & Tuned + ens. & 1{,}205.1 & $+44/-43$ & 49.84 & 0.419 & 20.6 & -- \\
NeuralNet (FastAI) & Tuned + ens. & 1{,}194.1 & $+49/-58$ & 50.84 & 0.407 & 20.5 & -- \\
RandomForest & Tuned + ens. & 1{,}172.7 & $+51/-42$ & 52.76 & 0.384 & 21.7 & -- \\
Nori-30M & Default & 1{,}159.2 & $+73/-82$ & 53.97 & 0.369 & 23.9 & 75\% \\
Nori & Default & 1{,}147.9 & $+69/-77$ & 54.96 & 0.358 & 24.2 & 75\% \\
TabSTAR & Tuned & 1{,}093.3 & $+74/-81$ & 59.59 & 0.303 & 26.3 & -- \\
OrionMSP & Default & 1{,}086.9 & $+50/-49$ & 60.10 & 0.296 & 24.5 & 25\% \\
PerpetualBooster & Tuned + ens. & 1{,}085.5 & $+43/-44$ & 60.22 & 0.295 & 27.2 & -- \\
TabFlex & Default & 1{,}011.5 & $+60/-70$ & 65.79 & 0.229 & 28.4 & 25\% \\
KNeighbors & Tuned + ens. & 993.8 & $+60/-82$ & 67.01 & 0.214 & 28.5 & -- \\
LinearModel & Tuned + ens. & 962.6 & $+59/-100$ & 69.03 & 0.190 & 34.4 & -- \\
\end{longtable}

\clearpage
\begin{longtable}{llrrrrrr}
\caption{Complete classification TabArena board (38 datasets, 594 units): all 41
methods, ranked by Elo.}
\label{tab:board_cls_full}\\
\toprule
Method & Regime & Elo & 95\% int. & Avg.\ rank & Win rate & Gap (\%) & Imp. \\
\midrule
\endfirsthead
\multicolumn{8}{l}{\itshape Classification board (continued from previous page)}\\
\toprule
Method & Regime & Elo & 95\% int. & Avg.\ rank & Win rate & Gap (\%) & Imp. \\
\midrule
\endhead
\midrule
\multicolumn{8}{r}{\itshape continued on next page}\\
\endfoot
\bottomrule
\endlastfoot
TabFM+ & Ensemble & 1{,}790.7 & $+117/-83$ & 8.42 & 0.909 & 6.2 & -- \\
AutoGluon 1.6 (noncommercial, 4h) & AutoML & 1{,}760.8 & $+84/-54$ & 9.52 & 0.896 & 9.4 & -- \\
\textbf{\mitra{}} & Default & 1{,}756.3 & $+120/-78$ & 9.69 & 0.894 & 9.0 & -- \\
TabFM & Default & 1{,}755.1 & $+125/-111$ & 9.74 & 0.893 & 6.3 & -- \\
EXAONE Tabular & Default & 1{,}750.8 & $+80/-59$ & 9.91 & 0.891 & 10.0 & -- \\
AutoGluon 1.6 (extreme, 4h) & AutoML & 1{,}711.1 & $+61/-44$ & 11.59 & 0.871 & 10.1 & -- \\
AutoGluon 1.5 (4h) & AutoML & 1{,}654.2 & $+78/-67$ & 14.39 & 0.837 & 10.6 & -- \\
TabPFN-3 & Default & 1{,}628.7 & $+74/-62$ & 15.80 & 0.819 & 12.9 & -- \\
TabPFN-2.6 & Default & 1{,}577.3 & $+57/-54$ & 18.94 & 0.781 & 14.2 & -- \\
TabICLv2 & Default & 1{,}570.8 & $+70/-64$ & 19.37 & 0.776 & 13.5 & -- \\
RealTabPFN-2.5 & Tuned + ens. & 1{,}558.5 & $+64/-65$ & 20.19 & 0.766 & 14.1 & -- \\
AutoGluon 1.4 (4h) & AutoML & 1{,}479.7 & $+62/-51$ & 25.98 & 0.695 & 16.1 & -- \\
RealMLP & Tuned + ens. & 1{,}462.3 & $+51/-38$ & 27.38 & 0.678 & 16.7 & -- \\
TabM & Tuned + ens. & 1{,}441.1 & $+52/-40$ & 29.12 & 0.657 & 17.6 & -- \\
TabDPT-Turbo & Default & 1{,}419.3 & $+71/-55$ & 30.97 & 0.634 & 18.1 & -- \\
LightGBM & Tuned + ens. & 1{,}408.8 & $+44/-31$ & 31.88 & 0.623 & 18.4 & -- \\
TabICL & Default & 1{,}406.3 & $+47/-54$ & 32.09 & 0.621 & 18.5 & 5\% \\
LimiX & Default & 1{,}395.6 & $+82/-73$ & 33.03 & 0.609 & 18.4 & -- \\
TabDPT & Tuned + ens. & 1{,}395.1 & $+70/-48$ & 33.07 & 0.609 & 18.2 & -- \\
CatBoost & Tuned + ens. & 1{,}393.7 & $+51/-44$ & 33.20 & 0.607 & 18.0 & -- \\
iLTM & Tuned + ens. & 1{,}387.5 & $+44/-40$ & 33.75 & 0.601 & 18.4 & -- \\
ChimeraBoost & Tuned + ens. & 1{,}367.5 & $+52/-55$ & 35.53 & 0.579 & 19.3 & -- \\
XGBoost & Tuned + ens. & 1{,}359.1 & $+41/-45$ & 36.29 & 0.570 & 19.1 & -- \\
BetaTabPFN & Default & 1{,}357.5 & $+48/-47$ & 36.44 & 0.568 & 20.4 & -- \\
TabPFNv2 & Tuned + ens. & 1{,}350.2 & $+84/-72$ & 37.10 & 0.560 & 20.4 & 32\% \\
Mitra-v1 & Default & 1{,}350.2 & $+77/-78$ & 37.10 & 0.560 & 20.6 & 32\% \\
ModernNCA & Tuned & 1{,}347.3 & $+51/-35$ & 37.36 & 0.557 & 19.4 & -- \\
TabSwift & Default & 1{,}338.2 & $+52/-54$ & 38.20 & 0.546 & 20.1 & -- \\
xRFM & Tuned + ens. & 1{,}315.9 & $+50/-46$ & 40.26 & 0.521 & 20.3 & -- \\
Explainable Boosting & Tuned + ens. & 1{,}299.1 & $+37/-35$ & 41.82 & 0.502 & 21.1 & -- \\
NeuralNet (PyTorch) & Tuned + ens. & 1{,}296.0 & $+41/-37$ & 42.10 & 0.499 & 20.3 & -- \\
SAP-RPT-OSS & Default & 1{,}291.3 & $+53/-70$ & 42.54 & 0.493 & 21.9 & -- \\
NeuralNet (FastAI) & Tuned + ens. & 1{,}251.5 & $+60/-66$ & 46.22 & 0.448 & 22.1 & -- \\
ExtraTrees & Tuned + ens. & 1{,}209.3 & $+54/-50$ & 50.05 & 0.402 & 23.1 & -- \\
RandomForest & Tuned + ens. & 1{,}182.1 & $+60/-62$ & 52.45 & 0.373 & 24.1 & -- \\
TabSTAR & Tuned & 1{,}138.9 & $+72/-94$ & 56.12 & 0.328 & 27.1 & -- \\
OrionMSP & Default & 1{,}116.9 & $+60/-64$ & 57.92 & 0.306 & 27.2 & -- \\
PerpetualBooster & Tuned + ens. & 1{,}064.4 & $+55/-59$ & 61.96 & 0.257 & 31.8 & -- \\
LinearModel & Tuned + ens. & 1{,}040.3 & $+85/-88$ & 63.68 & 0.236 & 33.2 & -- \\
KNeighbors & Tuned + ens. & 1{,}026.5 & $+71/-89$ & 64.63 & 0.224 & 31.0 & -- \\
TabFlex & Default & 1{,}015.9 & $+82/-97$ & 65.33 & 0.215 & 32.4 & -- \\
\end{longtable}

\clearpage
\begin{longtable}{llrrrrrr}
\caption{Complete regression TabArena board (13 datasets, 222 units): all 40
methods, ranked by Elo.}
\label{tab:board_reg_full}\\
\toprule
Method & Regime & Elo & 95\% int. & Avg.\ rank & Win rate & Gap (\%) & Imp. \\
\midrule
\endfirsthead
\multicolumn{8}{l}{\itshape Regression board (continued from previous page)}\\
\toprule
Method & Regime & Elo & 95\% int. & Avg.\ rank & Win rate & Gap (\%) & Imp. \\
\midrule
\endhead
\midrule
\multicolumn{8}{r}{\itshape continued on next page}\\
\endfoot
\bottomrule
\endlastfoot
AutoGluon 1.6 (noncommercial, 4h) & AutoML & 2{,}078.1 & $+208/-155$ & 4.99 & 0.951 & 1.7 & -- \\
TabFM+ & Ensemble & 2{,}078.0 & $+205/-146$ & 5.00 & 0.951 & 1.3 & -- \\
AutoGluon 1.6 (extreme, 4h) & AutoML & 2{,}015.5 & $+176/-115$ & 6.28 & 0.935 & 2.2 & -- \\
TabFM & Default & 1{,}986.8 & $+170/-96$ & 6.96 & 0.926 & 2.4 & -- \\
\textbf{\mitra{}} & Default & 1{,}985.6 & $+223/-134$ & 6.99 & 0.926 & 1.5 & -- \\
EXAONE Tabular & Default & 1{,}879.5 & $+137/-95$ & 10.07 & 0.888 & 3.5 & -- \\
TabPFN-3 & Default & 1{,}798.2 & $+212/-125$ & 13.04 & 0.851 & 3.2 & -- \\
AutoGluon 1.5 (4h) & AutoML & 1{,}773.5 & $+136/-88$ & 14.06 & 0.839 & 4.7 & -- \\
Nori-30M & Default & 1{,}756.6 & $+119/-78$ & 14.79 & 0.830 & 4.2 & -- \\
RealTabPFN-2.5 & Tuned + ens. & 1{,}730.3 & $+133/-92$ & 15.98 & 0.815 & 4.1 & -- \\
TabPFN-2.6 & Default & 1{,}729.7 & $+94/-50$ & 16.00 & 0.815 & 4.8 & -- \\
TabDPT & Tuned + ens. & 1{,}722.1 & $+156/-87$ & 16.36 & 0.810 & 5.0 & -- \\
TabICLv2 & Default & 1{,}682.4 & $+224/-138$ & 18.31 & 0.786 & 4.7 & -- \\
Nori & Default & 1{,}675.0 & $+128/-76$ & 18.69 & 0.782 & 5.3 & -- \\
RealMLP & Tuned + ens. & 1{,}644.8 & $+106/-60$ & 20.29 & 0.762 & 5.8 & -- \\
TabDPT-Turbo & Default & 1{,}603.0 & $+173/-93$ & 22.62 & 0.733 & 5.9 & -- \\
AutoGluon 1.4 (4h) & AutoML & 1{,}586.5 & $+95/-81$ & 23.58 & 0.721 & 7.1 & -- \\
ModernNCA & Tuned + ens. & 1{,}532.4 & $+132/-115$ & 26.84 & 0.681 & 8.0 & -- \\
CatBoost & Tuned + ens. & 1{,}481.6 & $+86/-60$ & 30.09 & 0.641 & 8.7 & -- \\
LightGBM & Tuned + ens. & 1{,}473.0 & $+78/-73$ & 30.67 & 0.634 & 9.1 & -- \\
xRFM & Tuned + ens. & 1{,}464.6 & $+100/-95$ & 31.23 & 0.627 & 8.2 & -- \\
TabM & Tuned + ens. & 1{,}447.8 & $+112/-77$ & 32.38 & 0.613 & 7.6 & -- \\
iLTM & Tuned + ens. & 1{,}434.9 & $+62/-44$ & 33.28 & 0.602 & 10.0 & -- \\
ChimeraBoost & Tuned + ens. & 1{,}407.8 & $+109/-104$ & 35.20 & 0.578 & 10.0 & -- \\
XGBoost & Tuned + ens. & 1{,}402.1 & $+54/-37$ & 35.61 & 0.573 & 9.6 & -- \\
TabSwift & Default & 1{,}397.7 & $+144/-108$ & 35.93 & 0.569 & 8.0 & -- \\
TabPFNv2 & Tuned + ens. & 1{,}322.4 & $+156/-116$ & 41.36 & 0.502 & 9.4 & 46\% \\
Mitra-v1 & Default & 1{,}265.2 & $+130/-109$ & 45.47 & 0.451 & 11.0 & 46\% \\
SAP-RPT-OSS & Default & 1{,}264.3 & $+154/-145$ & 45.54 & 0.450 & 11.5 & -- \\
LimiX & Default & 1{,}254.3 & $+169/-157$ & 46.25 & 0.441 & 11.0 & -- \\
NeuralNet (PyTorch) & Tuned + ens. & 1{,}235.1 & $+100/-105$ & 47.65 & 0.424 & 11.6 & -- \\
ExtraTrees & Tuned + ens. & 1{,}227.8 & $+107/-102$ & 48.18 & 0.418 & 13.5 & -- \\
PerpetualBooster & Tuned + ens. & 1{,}175.6 & $+67/-92$ & 51.90 & 0.372 & 13.8 & -- \\
RandomForest & Tuned + ens. & 1{,}159.1 & $+64/-76$ & 53.04 & 0.358 & 14.5 & -- \\
Explainable Boosting & Tuned + ens. & 1{,}144.6 & $+128/-186$ & 54.01 & 0.345 & 15.1 & -- \\
NeuralNet (FastAI) & Tuned + ens. & 1{,}024.1 & $+105/-117$ & 61.47 & 0.253 & 15.9 & -- \\
TabICL & Default & 1{,}000.0 & $+42/-75$ & 62.83 & 0.237 & 16.5 & 100\% \\
TabSTAR & Tuned + ens. & 963.9 & $+235/-313$ & 64.77 & 0.213 & 24.0 & -- \\
KNeighbors & Tuned + ens. & 886.0 & $+151/-166$ & 68.49 & 0.167 & 21.5 & -- \\
LinearModel & Tuned + ens. & 499.7 & $+128/-378$ & 78.75 & 0.040 & 37.8 & -- \\
\end{longtable}
\renewcommand{\arraystretch}{1.0}
\normalsize
\section{Negative results}
\label{app:negative}

During development we also tried several directions that did not work. None
improved on the released configuration, so we kept the simpler recipe. We group
the main negative results below so that others do not need to repeat them.

\paragraph{Regression objective and head.}
Alternatives to the released binned cross-entropy regression head did not improve
accuracy: HL-Gauss and CRPS fine-tuning objectives, a quantile head in place of
the binned one, and several variants of how targets are binned, embedded, and
decoded.

\paragraph{More or continued training.}
Training the released checkpoints further did not help. Warm-started continued
pretraining washed out, and simply training for longer left the board flat. We
also tried a late round of continued pretraining on priors that were rebalanced
toward the stronger feature correlations of real data. These runs used
correlation-matched copula-style and remixed variants, with and without EMA
weight averaging, and all of them lost ground across the regression suite.

\paragraph{Synthetic-prior settings.}
Two knobs on the synthetic prior turned out not to matter much: sampling the
number of support rows log-uniformly rather than uniformly, and reweighting the
outer prior mixture; both left downstream accuracy essentially unchanged
(\Cref{sec:priormix,sec:taskshapes}).

\paragraph{Deployment-time levers.}
Changes to how the frozen model is used at prediction time either made no
difference or hurt. One-hot expansion of categorical columns hurt accuracy, and
other feature selectors did no better than the released rules. The clearest case
was context length. We raised the classification support cap well past its
released value, using exact chunked attention so that memory was never the
limit, and accuracy still dropped: the model cannot use support sets much longer
than the ones it was fine-tuned on, so a few datasets get much worse. Raising the
fine-tuning support budget instead looked good on a single split, but on the
full multi-split evaluation the gain shrank to under one Elo; it is kept in the
released configuration only because it costs nothing. It cannot be combined with
a still larger prediction-time cap, because the two together run out of memory. Scaling
the logit temperature with support length gave no reliable gain, and picking the
support rows by stratified or clustering-based selection did not help and
sometimes hurt.

\paragraph{Per-dataset selection.}
Finally, choosing the configuration separately for each dataset does score
higher. The support cap in particular could be set per dataset to avoid the sharp
drops above. But this only works as an oracle: it amounts to tuning on the
benchmark. Every number in this report instead comes from one uniform
configuration, computable from the training data alone.
\section{2D visualizations of the Hybrid SCM prior}
\label{app:2D_visualization}

\begin{figure}[p]
  \centering
  \includegraphics[width=\textwidth]{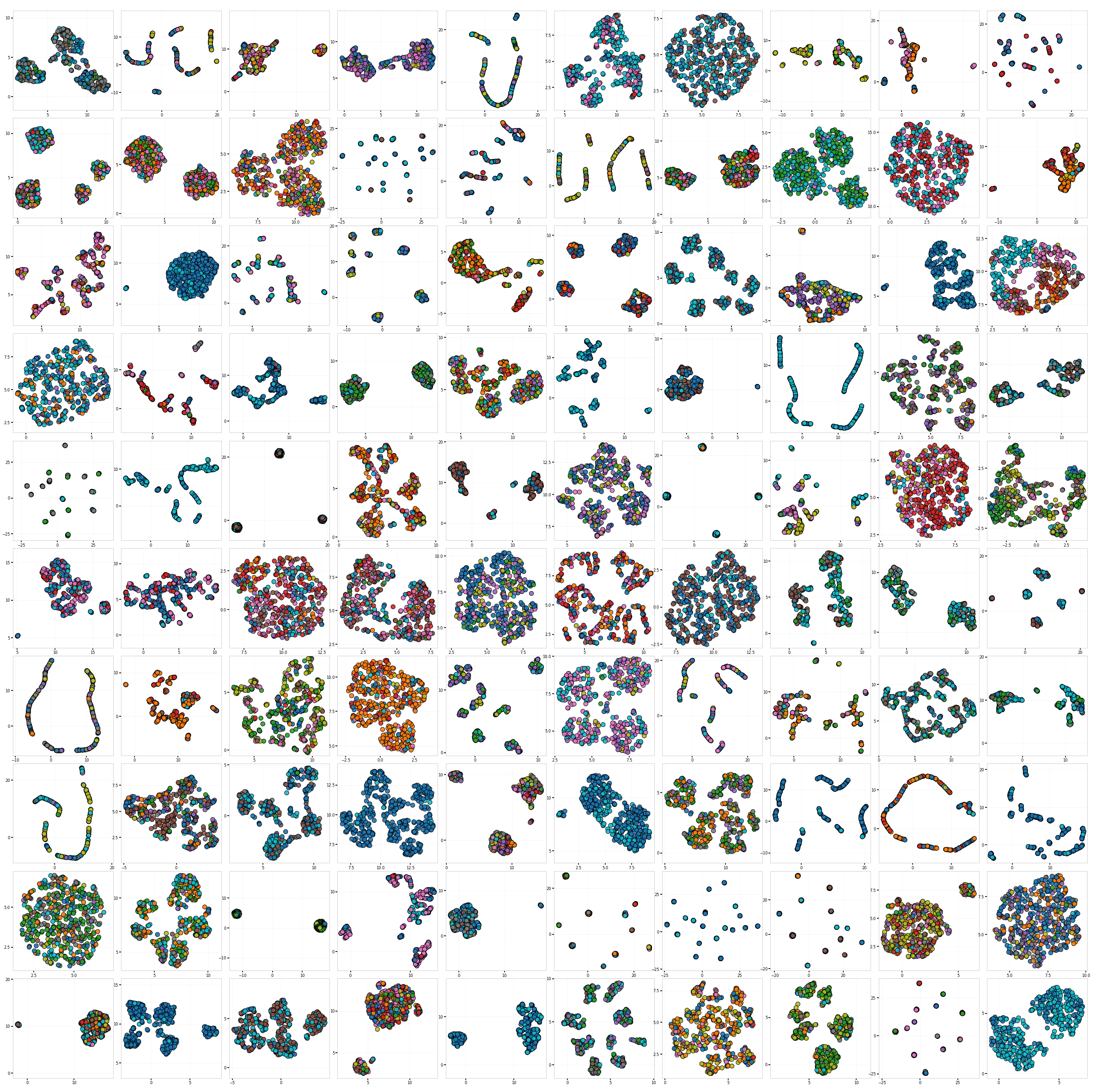}
  \caption{UMAP projections (after PCA whitening) of datasets sampled from the
Hybrid SCM data prior with $1$--$16$ feature dimensions, the range shared with
Mitra-v1. Each panel is one sampled dataset, with points colored by class label
$y$. The prior spans a broad range of geometric structures, including compact
clusters, rings, filaments, and curved manifolds.}
  \label{fig:umap-low}
\end{figure}

\begin{figure}[p]
  \centering
  \includegraphics[width=\textwidth]{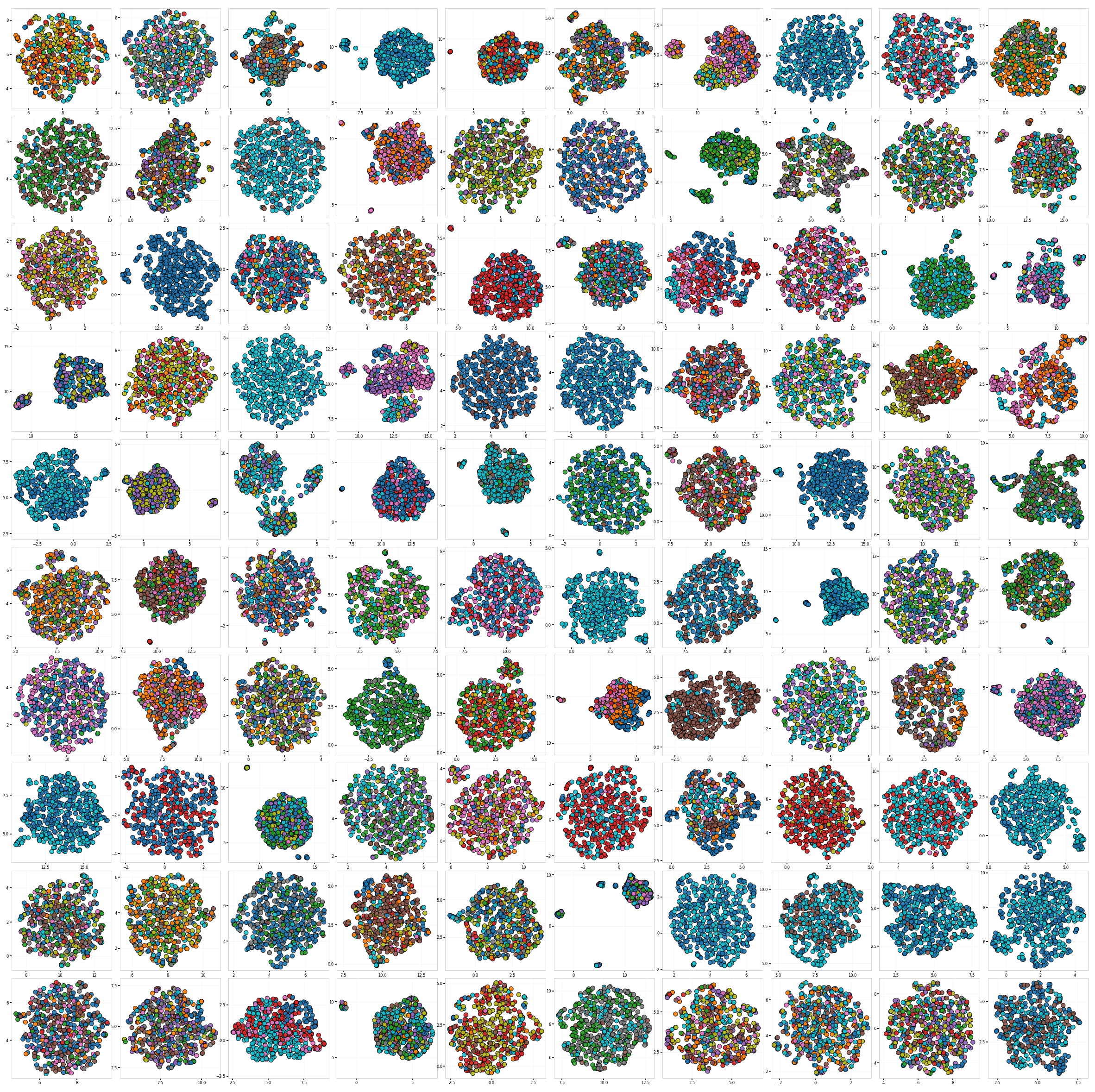}
  \caption{UMAP projections (after PCA whitening) of datasets sampled from the
Hybrid SCM data prior with $17$--$50$ feature dimensions, a higher-dimensional
setting that Mitra-v1 does not generate. Each panel is one sampled dataset, with
points colored by class label $y$. Compared with the $1$--$16$ range
(Figure~\ref{fig:umap-low}), these datasets are dominated by denser, blob-like
clusters.}
  \label{fig:umap-high}
\end{figure}

Figures~\ref{fig:umap-low} and~\ref{fig:umap-high} visualize datasets sampled from our Hybrid SCM prior: each small panel is one dataset reduced to two dimensions with UMAP after PCA whitening, with points colored by class label $y$. Figure~\ref{fig:umap-low} covers the 1--16 feature range shared with Mitra-v1; Figure~\ref{fig:umap-high} covers the 17--50 range, which Mitra-v1 does not generate. Together, the two figures show that the prior covers a wide span of both dataset geometry and feature dimensionality. Each figure is shown at full size on its own page for legibility.

\FloatBarrier
\section{Mitra use case overview}
\label{app:usecases}

Since the release of Mitra-v1 in July 2025, the Mitra family of models has been
downloaded more than 12 million times and applied and independently evaluated in
a growing number of published studies spanning energy, healthcare,
transportation, finance, and manufacturing. We list only studies that directly
ran Mitra.

\subsection{Healthcare and life sciences}
\begin{itemize}[leftmargin=1.2em,itemsep=4pt]
  \item Mitra was converted into a survival predictor by casting time-to-event
  modeling as a sequence of classification problems (MITRA-CF), achieving the
  best overall performance among fourteen survival models on 43 static clinical
  datasets and the best raw performance in dynamic prediction on 5 longitudinal
  datasets (integrated AUC 0.693 vs.\ 0.642 for the runner-up).
  \href{https://arxiv.org/abs/2601.22259}{Link}
  \item Mitra was evaluated as one of five tabular foundation model backbones for
  censoring-aware survival prediction with a pseudo-RMST regression target on
  SurvSet clinical benchmarks.
  \href{https://arxiv.org/abs/2607.09577}{Link}
\end{itemize}

\subsection{Financial services, banking, and insurance}
\begin{itemize}[leftmargin=1.2em,itemsep=4pt]
  \item Mitra was evaluated zero-shot for probability-of-default prediction
  across 14 credit-risk datasets and 29 methods, covering discrimination,
  calibration, and decision-oriented metrics relevant to capital allocation,
  provisioning, and pricing, reaching an average AUC of 0.724 without
  dataset-specific tuning.
  \href{https://arxiv.org/abs/2605.18147}{Link}
\end{itemize}

\subsection{Energy and utilities}
\begin{itemize}[leftmargin=1.2em,itemsep=4pt]
  \item Mitra, applied zero-shot through AutoGluon in a rolling-window setup, was
  the most accurate of four base forecasters for day-ahead electricity prices on
  the German EPEX and Spanish OMIE markets at every temporal aggregation level,
  with temporal hierarchy reconciliation improving accuracy further.
  \href{https://arxiv.org/abs/2508.11372}{Link}
\end{itemize}

\subsection{Transportation and mobility}
\begin{itemize}[leftmargin=1.2em,itemsep=4pt]
  \item Mitra delivered the most consistent and balanced crash injury-severity
  predictions among four tabular deep learning models on 18,073 glare-related
  Texas traffic records, reaching 93.3\% accuracy in a SHAP-based explainable
  safety analysis.
  \href{https://doi.org/10.1007/s42421-026-00163-7}{Link}
  \item Mitra's in-context predictions were embedded in a constrained
  multinomial-logit adapter for travel-demand and willingness-to-pay modeling,
  gaining 4 to 14 accuracy points over classical logit on Swissmetro, LPMC, and
  IoT-Wearables while restoring economic guarantees such as cost monotonicity.
  \href{https://arxiv.org/abs/2606.26432}{Link}
  \item Mitra, accessed through AutoGluon, raised travel-mode choice prediction
  accuracy well above multinomial logit (77.7\% vs.\ 63.7\% on Swissmetro), and a
  two-stage econometric adapter built on its predictions retained the gain while
  restoring valid value-of-time estimates for transport policy analysis.
  \href{https://arxiv.org/abs/2605.26559}{Link}
\end{itemize}

\subsection{Industrial, manufacturing, and supply chain}
\begin{itemize}[leftmargin=1.2em,itemsep=4pt]
  \item Mitra was evaluated zero-shot for supplier lead time estimation on real
  anonymized procurement datasets from Kinaxis supply chain deployments,
  outperforming TabICLv2 among the foundation-model baselines.
  \href{https://arxiv.org/abs/2607.18530}{Link}
\end{itemize}

\subsection{Retail, marketing, and customer analytics}
\begin{itemize}[leftmargin=1.2em,itemsep=4pt]
  \item Mitra was one of six tabular foundation models benchmarked for
  customer-churn prediction across nine public datasets spanning seven industry
  sectors, where foundation models collectively and decisively outperformed all
  classical ML, deep learning, and tree-ensemble alternatives.
  \href{https://openreview.net/forum?id=LtXucHLtiN}{Link}
\end{itemize}

\subsection{Real estate and geospatial}
\begin{itemize}[leftmargin=1.2em,itemsep=4pt]
  \item Mitra was evaluated zero-shot and fine-tuned via AutoGluon for estimating
  urban location values from rents in Dortmund, Germany, performing within a
  narrow margin of a stacked classical ensemble and producing spatially coherent
  value maps that local experts judged plausible.
  \href{https://doi.org/10.5194/agile-giss-7-15-2026}{Link}
\end{itemize}

\subsection{Science and engineering}
\begin{itemize}[leftmargin=1.2em,itemsep=4pt]
  \item Mitra was evaluated zero-shot as one of four tabular foundation model
  backbones for multi-fidelity regression on physics simulation (aerodynamics,
  CFD) and hyperparameter optimization benchmarks, outperforming the strongest
  Gaussian-process baseline.
  \href{https://arxiv.org/abs/2601.22371}{Link}
\end{itemize}

\subsection{Ecosystem and deployment integrations}
\begin{itemize}[leftmargin=1.2em,itemsep=4pt]
  \item Mitra ships natively in AutoGluon Tabular, supporting classification,
  regression, zero-shot inference, and fine-tuning through the standard AutoGluon
  interface.
  \href{https://auto.gluon.ai/stable/tutorials/tabular/tabular-foundational-models.html}{Link}
  \item A community DuckDB extension enables zero-shot Mitra classification and
  regression directly from SQL, with local ONNX inference on CPU, CUDA, and ROCm.
  \href{https://github.com/DataZooDE/anofox-tabfm}{Link}
  \item A community ONNX port enables Mitra inference from Node.js and browser
  environments.
  \href{https://github.com/wlearn-org/mitra-onnx}{Link}
\end{itemize}
\section{Automated checkpoint search with LLMZero}
\label{app:llmzero}

Empirically, validation Elo does not increase monotonically with pretraining steps, rendering the naive continuation of a given checkpoint an inefficient strategy for model improvement (\Cref{app:negative}). Consequently, we argue that checkpoint optimization is better formulated as a search over continuation trajectories rather than a selection along a single, static trajectory. We investigate this paradigm using LLMZero \citep{llmzero}, a post-training agent originally developed to discover reinforcement learning strategies, driven by the budget-constrained agentic search algorithm ExTS \citep{exts}. Here, we adapt this framework for continued pretraining: given an initial checkpoint, the agent launches short continuation segments, proposes trajectory-conditioned hyperparameter adjustments, and searches for an optimal continuation strategy.

Starting from a checkpoint of an independent pretraining run, a guided search of eight iterations discovered a 750-step continuation that improved validation Elo by $+17.1$. In contrast, unguided continuation along the original trajectory failed to yield improvements at any length. Naturally, the absolute performance of this method is bounded by its initialization. Because the search refines an existing checkpoint, i.e., resuming from its specific weights and optimizer states, the reachable performance is constrained by the quality of the source run. Consequently, while the improved checkpoint demonstrated significant relative gains, it did not surpass our primary released model, which originates from a strictly stronger pretraining run. Nevertheless, this controlled experiment provides a compelling proof of concept: automated, training-dynamics-guided trajectory search recovers substantial gains inaccessible to standard continuation. We leave the application of this search to our strongest available checkpoints for future work.

\end{document}